%% file: main.tex
\documentclass[accepted,hidelinks,onefignum,onetabnum]{siamart220329}

\input{shared}

\ifpdf
\hypersetup{
  pdftitle={DG Solutions via Small Linear CNNs},
  pdfauthor={A. Celaya, Y. Wang, D. Fuentes, B. Riviere}
}
\fi

\begin{document}

\maketitle

% REQUIRED
\begin{abstract}
In recent years, there has been an increasing interest in using deep learning and neural networks to tackle scientific problems, particularly in solving partial differential equations (PDEs). However, many neural network-based methods, such as physics-informed neural networks, depend on automatic differentiation and the sampling of collocation points, which can result in a lack of interpretability and lower accuracy compared to traditional numerical methods.  To address this issue, we propose two approaches for learning discontinuous Galerkin solutions to PDEs using small linear convolutional neural networks. Our first approach is supervised and depends on labeled data, while our second approach is unsupervised and does not rely on any training data. In both cases, our methods use substantially fewer parameters than similar numerics-based neural networks while also demonstrating comparable accuracy to the true and DG solutions for elliptic problems.
\end{abstract}

% REQUIRED

% REQUIRED

\section{Introduction}
Partial differential equations (PDEs) are commonly used to model physical phenomena and are thus widely used in many fields of science and engineering. Traditional numerical methods for approximating solutions to PDEs involve partitioning the computational domain in a mesh and discretizing the differential operators in finite dimensional spaces. This discretization leads to large linear systems, and efficiently solving these systems poses several computational challenges. Since their introduction in 2017, physics-informed neural networks (PINNs) have gained popularity as mesh-free alternatives to traditional numerical methods for solving PDEs \cite{raissi2017physics, raissi2017physics}. Thanks to the universal approximation property of neural networks, PINNs are suitable methods to  approximate the solution to a PDE \cite{Hornik, Funahashi, dissanayake1994neural, lagaris1998artificial, guhring2020, zeinhofer2024unified}. 

A crucial aspect of PINNs is the use of auto-differentiation to compute a residual-based loss function at a set of sampled collocation points. Although PINNs provide a mesh-free solution and have shown promise across various fields - such as biology \cite{zhang2023physics, li2024inverse}, meteorology \cite{kashinath2021physics, soto2024physics}, and optimal control \cite{mowlavi2023optimal, demo2023extended} - their reliance on auto-differentiation and sampling leads to challenges in interpretability and lower accuracy compared to traditional numerical methods \cite{chiu2022can, blechschmidt2021three, markidis2021old}. Choosing collocation points involves sampling points from the computational domain. However, the distribution and number of collocation points needed to accurately represent the solution of a given PDE remain unclear \cite{hou2023enhancing, wu2023comprehensive}, in particular for solutions with localized high gradients. Additionally, determining the best neural network architecture for the PINN itself is challenging \cite{wang2024pinn}. If the complexity of a PDE requires a large neural network, this can significantly increase the computational cost of training the network in terms of both time and memory \cite{stiasny2023physics}. 

The term numerics-informed neural network (NINN) was introduced in \cite{celaya2024solutions} to describe neural network-based approaches for solving PDEs that use traditional numerical methods (i.e., finite difference and finite element) to guide the design of the network architectures or loss functions, resulting in more explainable solutions than PINNs. In \cite{celaya2024solutions}, the authors use small linear convolutional neural networks (CNNs) to learn the finite difference solutions to elliptic and time-dependent parabolic problems. Similar approaches for learning finite difference solutions are proposed in \cite{Zhao2023, chiu2022can, Lim2022}. However, these methods inherit the limitations of the finite difference method, which requires smoothness of the solution for convergence.

The deep Ritz and deep Galerkin methods incorporate aspects of finite element methods, allowing for faster convergence during training \cite{SirignanoDGM:2018, WangDRM:2018, MullerZeinhofer}. However, the deep Ritz and Galerkin methods rely on feed-forward neural networks that take points sampled from the computational domain as inputs. The relationship between the input points is not explicitly considered by the network architecture, potentially resulting in a loss of valuable information. CNNs and graph neural networks (GNNs) address the limitations of feed-forward neural networks by directly incorporating spatial information into their underlying architectures. CNNs are used to learn finite element solutions in \cite{rezasefat2023finite}, where the authors use a U-Net-based CNN architecture trained on datasets generated through finite element simulations. Because of their ability to represent unstructured meshes, recent works have used GNNs to learn finite element methods. For example, \cite{gao2022physics} proposes a physics-informed GNN with a variational loss function to learn finite element solutions to several forward and inverse problems. Additional examples of GNN-based finite element solvers are investigated in \cite{gulakala2023graph,shivaditya2022graph, black2022learning}. However, like with the finite difference-based NINNs, these methods inherit the limitations of the finite element method. 

There has been an increased interest in incorporating more recent discretization methods like discontinuous Galerkin (DG) methods into the field of deep learning. It is known that
DG methods are locally mass conservative and are better suited for coupled flow and transport problems. They easily handle solutions with sharp gradients and discontinuities.\cite{hesthaven2007nodal, riviere2008discontinuous}. In the context of NINNs, \cite{li2024implementing} uses a combination of CNNs and GNNs to represent the different terms in the bilinear form of the symmetric interior penalty DG discretization to compute the PDE residual. Additionally, this approach requires that the user provides an initial guess for the linear system derived from the DG scheme. However, this method is not tested on different configurations of the DG scheme (i.e., different penalty parameters and symmetries) and does not specify the number of trainable parameters used in the neural network. Based on the peak memory usage presented in this approach, we estimate that the number of parameters used in their proposed architecture is roughly $10^7$. An alternative method is proposed in \cite{baharlouei2023dnn}, which uses a feed-forward network architecture to learn hybridized DG \cite{cockburn2009unified} solutions to elliptic problems. While this approach shows accurate results for its test cases, it is possible that this approach could benefit from more advanced network architectures like CNNs or GNNs. 

In this work, we propose two novel CNN-based approaches for learning DG solutions to elliptic problems. The first approach is supervised and uses labeled training data. It employs a loss function that calculates the $L^2$ and $H^1$ errors between the DG solutions and the predicted solutions, as well as the residual from the linear system stemming from the DG scheme. The second is unsupervised, meaning that it does not depend on labeled training data.  Instead, it takes advantage of the local mass conservation property of DG methods in the loss function. In both cases, we use small linear CNNs to achieve comparable accuracy to different configurations of the symmetric and non-symmetric interior penalty methods. Additionally, we show that the predicted solutions from both of our approaches are good initial guesses for iterative linear solvers, indicating that our approach can be used in conjunction with traditional DG solvers to improve their computational performance. 

An outline of the paper is as follows. Section~\ref{sec:methods} introduces the model problem, the DG scheme we use to generate data, the loss functions for the supervised and unsupervised approaches, and our network architecture. Results and discussion are presented in Section~\ref{sec:results}. Conclusions follow.

\section{Methods} \label{sec:methods}
%We propose two approaches for learning DG solutions to elliptic problems. The first is a supervised approach that uses labeled training data. 

% In this section, we define our model problem in Section \ref{sec:methods-model-problem}, followed by a description of the interior penalty DG method in Section \ref{sec:methods-dg-scheme}. Section \ref{sec:methods-data} outlines how we generate data to train our methods. Sections \ref{sec:methods-supervised} and \ref{sec:methods-unsupervised} detail our supervised and unsupervised approaches for estimating DG solutions, respectively. Finally, Section \ref{sec:methods-training} explains our training and testing protocols.

\subsection{Model problem} \label{sec:methods-model-problem}
Let $\Omega \subset \mathbb{R}^2$ be an open domain with Lipschitz boundary. Given a function $f\in L^2(\Omega)$, the function $u\in H^1(\Omega)$ is the solution to the Poisson problem with homogeneous Dirichlet boundary condition:
\begin{align*}
    \begin{split}
    -\Delta u & = f, \quad \text{in } \Omega,\\
    u &= 0, \quad \text{on } \partial \Omega.
    \end{split}
\end{align*}
For simplicity, we assume that $\Omega$ is the square domain $(0,L)^2$. The proposed methods are applicable to rectangular domains partitioned into structured grids. 

\subsection{DG scheme} \label{sec:methods-dg-scheme}
Samples to train and test our neural networks are generated using the interior penalty DG method. The domain $\Omega$ is partitioned into a mesh, $\mathcal{E}_h$,  made of $M_h$ square elements $E$ of side $h\sqrt{2}$. We recall below the DG method of order one. Let $V_h$ be the space of discontinuous piecewise linear polynomials. The DG solution $u_{h}\in V_h$ satisfies the discrete variational problem
\begin{equation}\label{eq:dgpb}
a(u_{h},v_h) = \ell(v_h), \quad \forall v_h\in V_h,
\end{equation}
where the bilinear form $a(\cdot,\cdot)$ and the form $\ell$ are such that for any $w_h, v_h$ in $V_h$:
\begin{align*}
    a(w_h,v_h) = &\sum_{E \in \mathcal{E}_h} \int_E \nabla w_h \cdot \nabla v_h \\ 
    -&\sum_{\gamma \in \Gamma_h} \int_\gamma \{\nabla w_h  \cdot \mathbf{n}_\gamma\}[v_h] 
    + \varepsilon \sum_{\gamma \in \Gamma_h} \int_\gamma \{\nabla v_h \cdot \mathbf{n}_\gamma\} [ w_h] \\%& \text{interior edges terms}\\
    - &\sum_{\gamma \in \partial \Omega} \int_\gamma \nabla w_h \cdot \mathbf{n} \, v_h
    + \varepsilon \sum_{\gamma \in \partial \Omega} \int_\gamma \nabla v_h \cdot \mathbf{n} \, w_h \\%& \text{boundary terms}\\
    + &\sum_{\gamma \in \Gamma_h} \frac{\sigma}{h} \int_\gamma [w_h][v_h]
    +\sum_{\gamma \in \partial \Omega} \frac{\sigma}{h} \int_\gamma w_h v_h,\\
    \ell(v_h) = &\int_\Omega f v_h.
\end{align*}
The parameter $\varepsilon$ takes the value $+1$ or $-1$, which yields the non-symmetric interior penalty Galerkin (NIPG) method or the symmetric interior penalty Galerkin (SIPG) method respectively. The penalty parameter $\sigma$ is a positive constant that is chosen large enough if $\varepsilon = -1$. 

To complete the definitions above, we introduce some notation.  The outward unit normal vector to $\partial\Omega$ is denoted by $\mathbf{n}$. The set of interior edges is denoted by $\Gamma_h$. On each interior edge $\gamma$, we fix a unit normal vector $\mathbf{n}_\gamma$ and we denote by $E_{1,\gamma}$ and $E_{2,\gamma}$ the neighboring elements of $\gamma$ such that $\mathbf{n}_\gamma$ points from $E_{1,\gamma}$ into $E_{2,\gamma}$. With this convention, the jump of a discontinuous function $v_h$ across the edge $\gamma$ is defined as:
\[
[v_h]|_\gamma =  v_h|_{E_{1,\gamma}} - v_h|_{E_{2,\gamma}}.
\]
The average $\{ v_h \}$ is simply the arithmetic average of the traces:
\[
\{v_h\}|_\gamma =  \frac12 v_h|_{E_{1,\gamma}} + \frac12 v_h|_{E_{2,\gamma}}.
\]
The restriction of $u_{h}$ to each element $E$ is written as
\begin{align*}
u_{h}(x,y) \rvert _E = \hat{u}_{h}(\hat x,\hat y) = \sum_{j=0}^3 \alpha_{j,E} \psi_j(\hat{x},\hat{y}),
\end{align*}
where $(\hat{x},\hat{y})\mapsto (x,y)$ is the affine transformation from the reference element $(-1,1)^2$ to the element $E$; the local degrees of freedom are $\bfalpha_E = (\alpha_{i,E})_{0\leq i\leq 3}$ and the local basis functions are:
\begin{align*}
    \psi_0(\hat x,\hat y) = \frac{1}{4} (1-\hat{x})(1-\hat{y}), \quad
    \psi_1(\hat x,\hat y) = \frac{1}{4} (1+\hat{x})(1-\hat{y}),\\
    \psi_2(\hat x,\hat y) =  \frac{1}{4} (1+\hat{x})(1+\hat{y}), \quad
    \psi_3(\hat x,\hat y) = \frac{1}{4} (1-\hat{x})(1+\hat{y}). 
\end{align*}
%The vector of unknowns $\bfalpha$ consists of small block vectors  $\bfalpha_E$.  

Taking advantage of the locality of the DG basis, the DG solution satisfies an element-wise mass error (obtained by choosing in \eqref{eq:dgpb} a test function $v_h=1$ on one element $E$ and $v_h=0$ elsewhere).  
\begin{align} \label{eqn:local-mass}
0 =  \ell_{\mathrm{mass},E}^{\sigma}(u_h) = -\int_{\partial E} \{ \nabla u_h \cdot \mathbf{n}_E\}
+\frac{\sigma}{h} \sum_{\gamma\subset\partial E} \int_\gamma ({u_h}|_E
-{u_h}|_{E^\gamma})
-\int_E f, \quad \forall E\in\mathcal{E}_h.
\end{align}
Here, the unit vector $\mathbf{n}_E$ is the normal outward of $\partial E$, and ${u_h}|_{E^\gamma}$ is the restriction of $u_h$ to each neighbor $E^\gamma$ across a face $\gamma$.

\subsection{Data} \label{sec:methods-data}
Our supervised and unsupervised approaches use CNNs that take as input the $L^2$ projection, $\pi f$,  of a source function $f$ onto the mesh:
\[
\forall v_h\in V_h, \quad \forall E \in \mathcal{E}_h, \quad \int_E (\pi f) \, v_h = \int_E f \, v_h.
\]
Using the same basis functions as above, we associate with the piecewise linear function $\pi f$, the vector $\bfpi_f\in\mathbb{R}^{4 M_h}$. 
Problem~\eqref{eq:dgpb} is equivalent to the square  linear system:
\begin{equation}\label{eq:linearsystem}
\mathbf{A}\bfalpha(f) = \bfpi_f,
\end{equation}
where  
$\bfalpha(f)\in\mathbb{R}^{4 M_h}$ is the vector of unknowns
that uniquely defines the solution $u_h(f)\in V_h$.

To make $\bfpi_f$ compatible with CNN architectures, we reshape $\bfpi_f$ into a $2\sqrt{M_h} \times 2\sqrt{M_h}$ array. This reshape operation is such that it preserves the global and local spatial organization of the degrees of freedom.

Let $\mathcal{F}$ denote the set of given source functions. For our supervised approach, the training data consists of $\{ (\bfpi_f, \bfalpha(f)): \, f \in \mathcal{F}\}$. Let $\mathcal{S}$ be a set of functions that we refer to as a symbol bank, $\mathcal{G} = \{+, -, \times\}$ be a set of binary operators, and $\mathcal{B}$ be a set of bubble functions. To generate each element $f \in \mathcal{F}$, we first construct functions $u$ by taking a random number of symbols from $\mathcal{S}$, concatenating them with random operators from $\mathcal{G}$, and multiplying the resulting function by a random bubble function in $\mathcal{B}$. We remark that by construction, the function  $u$ satisfies the Dirichlet boundary condition. We then scale the resulting function $u$ such that its range is roughly between negative one and one. The final step in this process is to compute $f = -\Delta u$, which we do with Matlab's symbolic toolbox. 

For our numerical results below, the set of bubble functions is
\begin{align*}
    \mathcal{B} = \{&(x - x^2)(y - y^2), (x^3-x^2)(y^3 - y^2), 
    \sin(\pi x)\sin(\pi y), \sin(2\pi x)\sin(2\pi y)\}.
\end{align*} 
The training data uses the following symbol bank
\begin{align*}
\mathcal{S}_{\text{train}} = \{
&\cos(\pi x), \sin(\pi x), e^x, \sqrt{x + 1}, \\ 
&\cos(\pi y), \sin(\pi y), e^y, \sqrt{y + 1}, \\
&\cos(\pi xy), \sin(\pi xy), e^{xy}, \sqrt{xy + 1}, \\
&\cos(\pi(x + y)), \sin(\pi(x + y)), e^{x + y}, \sqrt{x + y + 1}, \\
&x^2, y^2, x^2 y^2, xy, xy^2, x^2 y, \\
&x^3, y^3, x^3 y, x^3 y^2, y^3 x, y^3 x^2, \\
&e^{x^2 + y^2}, \sin(\pi(x^2 + y^2)), \cos(\pi(x^2 + y^2))
\}.
\end{align*}

To assess how well our models generalize to new data, we use a different symbol bank $\mathcal{S}_{\text{test}}$ to generate a test set of $100$ functions. This second symbol bank is
\begin{align*} \label{eqn:test-symbols}
    \mathcal{S}_{\text{test}} = \{
    &\cos(2\pi x), \sin(2 \pi x), e^{-x}, \log(\sin(\pi x) + 1),  \\
    &\cos(2\pi y), \sin(2\pi y), e^{-y}, \log(\sin(\pi y) + 1), \\
    &\log(\sin(\pi x)\sin(\pi y^2) + 1), \log(\sin(\pi x^2)\sin(\pi y) + 1), \\
    &\cos(2\pi x y), \sin(2\pi x y), x^4, y^4
    \}
\end{align*}

We implement the random function generation described above and a solver for the SIPG and NIPG  methods in Matlab and use them to generate training and testing data for our CNNs. 

\subsection{Network architecture} \label{sec:methods-architecture}
We select as our architecture the widely used U-Net \cite{unet}. The U-Net was initially developed for image segmentation but is successful in several scientific machine learning tasks \cite{gravity1, gravity2, joint-inversion}. Let $\mathcal{N}_\theta$ denote the neural network, with $\theta$ representing the set of learnable parameters. The architecture of $\mathcal{N}_\theta$ consists of convolution block operators (blocks), a downsampling (or encoding) path by the use of bilinear interpolation, and an upsampling (or decoding) path by the use of average pooling. A channel-wise concatenation operation links each layer in the downsampling and upsampling paths. Each block consists of two 2D convolution layers. As in \cite{celaya2024solutions}, we omit the use of nonlinear activation functions, normalization layers, and bias terms, resulting in a linear network. Additionally, we use the PocketNet approach proposed in \cite{celaya2022pocketnet} and leave the number of feature maps (channels) at each resolution within our architecture constant. Unless otherwise mentioned, the number of feature maps at each resolution is equal to $32$. We define the depth, $d$, of each network as the number of downsampling operations present in the network. The size of the feature tensor is halved from depth $d$ to depth $d+1$. For our numerical results, the depth of the network depends on the size of the input. Namely, for a given input of size $M \times M$ and kernel size $k$, the depth is given by $d = \lceil \log_2(M / (k + 1)) \rceil$. This formula for $d$ ensures that the feature maps at the coarsest resolution are not smaller than the convolutional kernels. Additionally, dynamically defining $d$ ensures that for a given input, the network will see the full range of fine and coarse scale features. 

\begin{figure}[ht!]
    \centering
    \includegraphics[width=\linewidth]{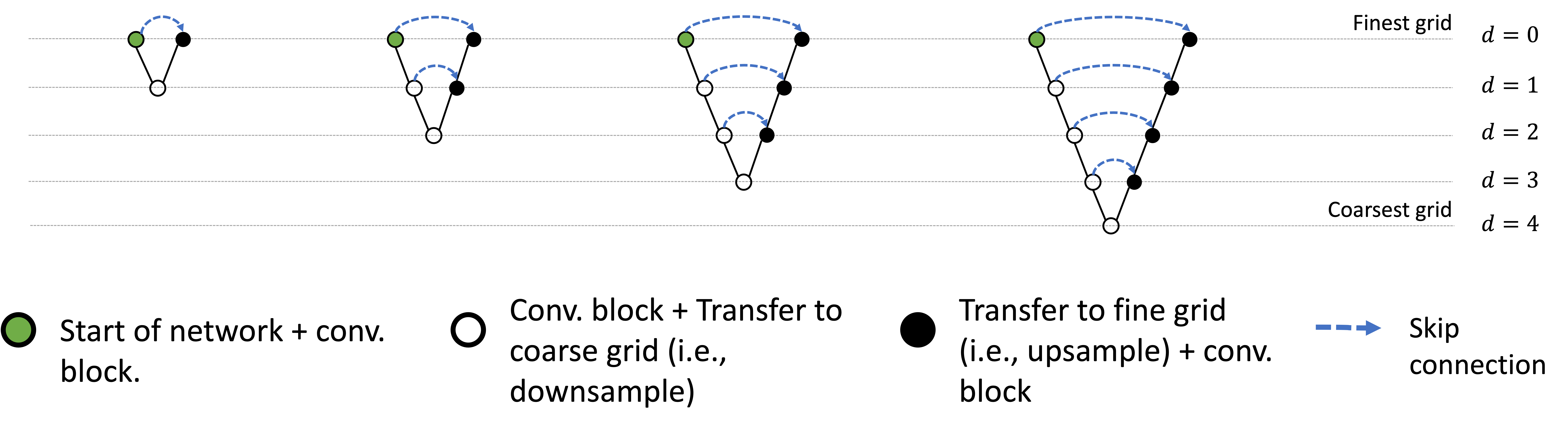}
    \caption{Sketch of U-Net architecture for different network depths: $0\leq d\leq 4$. From \cite{celaya2024solutions}.  \label{fig:network}}
\end{figure}

\subsection{Supervised approach for learning DG solutions to elliptic problems} \label{sec:methods-supervised}
We propose a supervised approach for learning DG solutions for elliptic problems that use the dataset and network architecture described in Sections \ref{sec:methods-data} and \ref{sec:methods-architecture}. This method uses labeled training data of the form $\mathcal{T} = \{ (\bfpi_f, \bfalpha(f)): \, f \in \mathcal{F}\}$ 
(with cardinality $|\mathcal{T}|$) and the following loss function:
\begin{align}
    \mathcal{L}_{\text{sup}}(u_h, \hat{u}_h) = \beta \left( \Vert u_h - \hat{u}_h \Vert_{L^2(\Omega)}^2 + |u_h - \hat{u}_h|_{H^1(\mathcal{E}_h)}^2\right) + (1 - \beta) \Vert \mathbf{A} \hat{\bfalpha}(f) - \bfpi_f \Vert_2^2,
\end{align}
where $\hat{u}_h$ is the predicted DG solution with corresponding degrees of freedom  $\hat{\bfalpha}(f)$,  and $\beta \in [0, 1]$ is a user-defined weighting parameter. By default, we set $\beta = 0.5$. For any piecewise function $v$, the broken $H^1$ seminorm is defined as 
\[
\vert v \vert_{H^1(\mathcal{E}_h)} = (\sum_{E\in\mathcal{E}_h} \Vert \nabla v \Vert_{L^2(E)}^2)^{1/2}.
\]
We note that the integrals in the $L^2$ norm and $H^1$ seminorm in the loss function are computed exactly with quadrature points defined on the reference element. 

Given a CNN $\mathcal{N}_{\theta}$, the supervised approach aims to solve the following minimization problem:
\begin{align*}
    \min_{\theta} \frac{1}{|\mathcal{T}|} \sum_{(\bfpi_f, \bfalpha(f)) \in \mathcal{T}} \mathcal{L}_{\text{sup}}(u_h, \mathcal{N}_{\theta}(\bfpi_f)).
\end{align*}
To start the algorithm, the learnable parameters (weights) are randomly initialized. At each iteration, backpropagation is applied to update them.

\subsection{Unsupervised approach for learning DG solutions to elliptic problems} \label{sec:methods-unsupervised}
In addition to our supervised approach for learning DG solutions to elliptic problems, we also propose an alternative unsupervised approach. Unlike the supervised approach described above, this method does not depend on labeled training data. Instead, it only uses the $L^2$ projection of the source function $\bfpi_f$ and takes advantage of the local mass conservation property defined in (\ref{eqn:local-mass}). We use the following loss function that minimizes the local mass error and the discontinuity of the predicted solution across interior mesh edges:
\begin{align}
    \mathcal{L}_{\text{uns}}(\hat{u}_h) &= \beta\sum_{E\in\mathcal{E}_h} (\ell_{\mathrm{mass},E}^{\sigma}(\hat u_h))^2
    +(1 - \beta)\eta \sum_{\gamma\in\Gamma_h} \Vert [\hat u_h]\Vert_{L^2(\gamma)}^2,
\end{align}
where $\beta \in [0,1]$ is the same weighting parameter that we use in the supervised approach, $\sigma > 0$ is a positive constant, and $\eta > 0$ is a positive constant that we refer to as the discontinuity penalty parameter. For this approach, we select the following default values: $\beta = 0.5$, $\sigma = 1$, and $\eta = 2$. The integrals in the definition of the unsupervised loss functions are computed using quadrature rule on the reference element. 

For a given $L^2$ projection $\bfpi_f$, our unsupervised approach solves the following minimization problem
\begin{align} \label{eqn:unsup}
    \min_{\theta} \mathcal{L}_{\text{uns}}(\mathcal{N}_{\theta}(\bfpi_f))
\end{align}
using gradient-based methods, where the neural network parameters $\theta$ are iteratively updated to minimize the unsupervised loss $\mathcal{L}_{\text{uns}}$. Like with the supervised method, the algorithm begins with a random initialization of the neural network, computes predictions based on the input $\bfpi_f$, evaluates the loss, and applies backpropagation to update the weights $\theta$ at each iteration. Throughout the training process, we track the best loss value and the corresponding best prediction $\hat{u}_h^*$, preserving the optimal solution. The detailed steps of the unsupervised training procedure are outlined in Algorithm \ref{alg:alg-unsup}.

\begin{algorithm}
\caption{
Unsupervised CNN training for (\ref{eqn:unsup}) \label{alg:alg-unsup} \\
 \textit{Input}: $L^2$ projection of source function $\bfpi_f$ \\
 \textit{Output}: Approximate solution to (\ref{eqn:unsup}), $\hat{u}_h^*$
 }
\begin{algorithmic}
\STATE{Randomly initialize neural network $\mathcal{N}_{\theta}$}
\STATE{Set maximum iterations $\mathcal{M}$}
\STATE{Start iteration counter: $k \leftarrow 0$}
\STATE{Store lowest loss value: $\ell_{*} \leftarrow \infty$}
\WHILE{$k < \mathcal{M}$}
\STATE{Get network prediction: $\hat{u}_h \gets \mathcal{N}_{\theta}(\bfpi_f)$}
\STATE{Compute loss: $\ell \gets \mathcal{L}_{\text{uns}}(\hat{u}_h)$}
\STATE{Update $\theta$ using backpropagation on $\ell$}
\IF{$\ell < \ell_*$}
\STATE{Save best prediction: $\hat{u}_h^* \gets \hat{u}_h$}
\STATE{Update smallest loss value: $\ell_* \gets \ell$}
\ENDIF
\STATE{Update iteration counter: $k \leftarrow k + 1$}
\ENDWHILE
\RETURN{$\hat{u}_h^*$}
\end{algorithmic}
\end{algorithm}

\subsection{Training and testing protocols} \label{sec:methods-training}
We train each model using the Adam optimizer with a cosine learning rate schedule. The initial learning rate is equal to $10^{-3}$. Throughout the training process, we apply $L^2$ regularization with a penalty of $10^{-7}$, and we clip the norm of the network's gradient to ensure it does not exceed $10^{-3}$. Our default settings for our network architecture are to use $32$ channels for each convolutional layer and $7\times 7$ kernels. In our supervised approach, the batch size is either $32$ or the total number of training examples, whichever is smaller, and we train each model for $150$ epochs. The number of functions in the training set is either fixed to $2000$ or it varies and takes the value  $10^m$ for $0\leq m\leq 3$. 
For our unsupervised approach, we set the default number of optimization steps $\mathcal{M} = 5,000$.
The set of $100$ test functions is fixed for all the numerical examples.

To assess the quality of our predictions, we use the $L^2$ norm and broken $H^1$ seminorm between either the true solution $u$ and the predicted DG solution $\hat{u}_h$ or between the predicted DG solution and the actual DG solution $u_h$. We remark that the CNN prediction $\hat u_h$ also belongs to  the DG space $V_h$.  Our models and loss functions are implemented in Python using PyTorch (v.2.5.1) and trained on an NVIDIA A100 GPU.

\section{Results and discussion} \label{sec:results}
We train models utilizing the methods outlined in Section \ref{sec:methods}, applying both our proposed supervised and unsupervised approaches. 

First, for a baseline, Table \ref{tab:results-dg-errors} shows the mean, standard deviation (numbers in parentheses), median, and convergence rate (with respect to the median) of the $L^2$  and broken $H^1$ errors between the exact
solution $u$ and the DG solution $u_h$ for $100$ test functions described in Section~\ref{sec:methods-training}. 
We vary the grid size, $h= \sqrt{2}/N$ with $N=16, 32, 64$. 
This means that the initial mesh contains $M_h = 16^2$ and it is successively refine to containt $32^2$ and  $64^2$ elements. We consider both SIPG and NIPG versions and vary the values of $\sigma$.  Convergence rates for the median errors are computed with respect to the grid size and as expected, we observe a rate equal to $1$ for the broken $H^1$ seminorm and a rate equal to $2$ for the $L^2$ norm.

\begin{table}[ht!]
\centering
\caption{Mean, standard deviation, median, and rate of convergence for the broken $H^1$ and $L^2$ errors between the true and DG solutions $u$ and $u_h$ for the SIPG and NIPG methods with varying values of $\sigma$.\label{tab:results-dg-errors}}
\bgroup
\def\arraystretch{1.75}%  1 is the default, change whatever you need
\resizebox{\textwidth}{!}{%
\begin{tabular}{lclccl|ccc|ccc}
\hline
\multicolumn{1}{c}{\multirow{2}{*}{Error}} & \multirow{2}{*}{Method} & \multirow{2}{*}{$\sigma$} & \multicolumn{3}{c|}{$N = 16$} & \multicolumn{3}{c|}{$N = 32$} & \multicolumn{3}{c}{$N = 64$} \\ \cline{4-12} 
\multicolumn{1}{c}{} &  &  & Mean (Std) & \multicolumn{2}{c|}{Median} & Mean (Std) & Median & Rate & Mean (Std) & Median & Rate \\ \hline
\multirow{6}{*}{$H^1$} & \multirow{3}{*}{SIPG} & 1 & 5.27e-01 (1.87e-01) & \multicolumn{2}{c|}{4.76e-01} & 2.38e-01 (8.37e-02) & 2.15e-01 & 1.14 & 1.08e-01 (3.68e-02) & 9.86e-02 & 1.13 \\
 &  & 5 & 3.64e-01 (1.21e-01) & \multicolumn{2}{c|}{3.26e-01} & 1.83e-01 (6.10e-02) & 1.64e-01 & 0.99 & 9.16e-02 (3.05e-02) & 8.22e-02 & 1.00 \\
 &  & 10 & 3.64e-01 (1.21e-01) & \multicolumn{2}{c|}{3.26e-01} & 1.83e-01 (6.10e-02) & 1.64e-01 & 0.99 & 9.15e-02 (3.05e-02) & 8.20e-02 & 1.00 \\ \cline{2-12} 
 & \multirow{3}{*}{NIPG} & 1 & 3.76e-01 (1.25e-01) & \multicolumn{2}{c|}{3.38e-01} & 1.86e-01 (6.20e-02) & 1.67e-01 & 1.01 & 9.24e-02 (3.08e-02) & 8.32e-02 & 1.00 \\
 &  & 5 & 3.64e-01 (1.21e-01) & \multicolumn{2}{c|}{3.26e-01} & 1.83e-01 (6.10e-02) & 1.64e-01 & 0.99 & 9.16e-02 (3.05e-02) & 8.22e-02 & 1.00 \\
 &  & 10 & 3.64e-01 (1.21e-01) & \multicolumn{2}{c|}{3.26e-01} & 1.83e-01 (6.10e-02) & 1.64e-01 & 0.99 & 9.15e-02 (3.05e-02) & 8.20e-02 & 1.00 \\ \hline
\multirow{6}{*}{$L^2$} & \multirow{3}{*}{SIPG} & 1 & 9.71e-03 (3.79e-03) & \multicolumn{2}{c|}{8.83e-03} & 2.14e-03 (7.94e-04) & 1.99e-03 & 2.15 & 4.66e-04 (1.63e-04) & 4.35e-04 & 2.19 \\
 &  & 5 & 5.17e-03 (1.70e-03) & \multicolumn{2}{c|}{1.62e-03} & 1.37e-03 (4.35e-04) & 1.25e-03 & 1.92 & 3.52e-04 (1.12e-04) & 3.24e-04 & 1.94 \\
 &  & 10 & 5.43e-03 (1.71e-03) & \multicolumn{2}{c|}{4.95e-03} & 1.40e-03 (4.45e-04) & 1.29e-03 & 1.93 & 3.56e-04 (1.13e-04) & 3.28e-04 & 1.98 \\ \cline{2-12} 
 & \multirow{3}{*}{NIPG} & 1 & 8.93e-03 (4.70e-03) & \multicolumn{2}{c|}{8.05e-03} & 2.38e-03 (1.27e-03) & 2.10e-03 & 1.94 & 6.15e-04 (3.29e-04) & 5.32e-04 & 1.98 \\
 &  & 5 & 4.11e-03 (1.53e-03) & \multicolumn{2}{c|}{3.80e-03} & 1.04e-03 (3.94e-04) & 9.61e-04 & 1.98 & 2.61e-04 (1.00e-04) & 2.41e-04 & 1.99 \\
 &  & 10 & 4.60e-03 (1.52e-03) & \multicolumn{2}{c|}{4.29e-03} & 1.14e-03 (3.79e-04) & 1.05e-03 & 2.02 & 2.85e-04 (9.46e-05) & 2.63e-04 & 2.00 \\ \hline
\end{tabular}%
}
\egroup
\end{table}

\subsection{Supervised approach}
Here, we present and discuss the results of our proposed supervised approach. Recall that supervised training uses labeled data. We first compare the accuracy of our supervised approach to the baseline accuracy shown in Table \ref{tab:results-dg-errors}. In this case, we fix the symmetrization parameter $\varepsilon$, the mesh size $h$ and
the penalty value $\sigma$. We then train our proposed CNN for
these fixed values. Once training is done ($150$ epochs), we apply the neural network to the set of $100$ test functions. Table \ref{tab:results-sup-sigma-n} shows the mean, standard deviation, and median $L^2$ and broken $H^1$ errors between the actual solution $u$ and the predicted CNN solution $\hat{u}_h$ for the 100 test cases. We also compute the rate of convergence for our predicted DG solutions using the median errors for each grid size. We see that our models achieve similar $H^1$ errors and convergence rates to those obtained with the DG solutions. We also observe that varying $\varepsilon$ or $\sigma$ does not have a significant impact on the accuracy or convergence rate.

Regarding the $L^2$ norm, we observe that the errors for the 16$\times$16 grid align closely with those of the DG solution. However, for higher resolution grids, we see that while the $L^2$ errors remain within the same order of magnitude as those obtained with the DG solutions, there is a loss of convergence rate as the grid is refined.  One possible explanation for this is that for higher resolutions, the Adam optimizer struggles to navigate the increasingly complex loss landscape, leading to suboptimal convergence. This issue may be compounded by the higher dimensionality of the solution space, which demands more training iterations or different learning rate schedules to resolve finer-scale errors. Exploring alternative optimization strategies, such as second-order methods or cascaded training, could mitigate these challenges and improve performance for finer grids.
We note that the convergence rate is larger if the penalty parameter $\sigma$ is set equal to $1$.

\begin{table}[ht!]
\centering
\caption{Mean, standard deviation, median, and rate of convergence for the broken $H^1$ and $L^2$ errors between the true solution $u$ and the predicted CNN solution $\hat{u}_h$ for $\varepsilon = -1, +1$ and for different values of $\sigma$.
\label{tab:results-sup-sigma-n}}
\bgroup
\def\arraystretch{1.75}%  1 is the default, change whatever you need
\resizebox{\textwidth}{!}{%
\begin{tabular}{lclccl|ccc|ccc}
\hline
\multirow{2}{*}{Error} & \multirow{2}{*}{Method} & \multirow{2}{*}{$\sigma$} & \multicolumn{3}{c|}{$N = 16$} & \multicolumn{3}{c|}{$N = 32$} & \multicolumn{3}{c}{$N = 64$} \\ \cline{4-12} 
 &  &  & Mean (Std) & \multicolumn{2}{c|}{Median} & Mean (Std) & Median & Rate & Mean (Std) & Median & Rate \\ \hline
\multirow{6}{*}{$H^1$} & \multirow{3}{*}{SIPG} & 1 & 5.26e-01 (1.87e-01) & \multicolumn{2}{c|}{4.76e-01} & 2.37e-01 (8.36e-02) & 2.15e-01 & 1.14 & 1.08e-01 (3.68e-02) & 9.86e-02 & 1.13 \\
 &  & 5 & 3.64e-01 (1.21e-01) & \multicolumn{2}{c|}{3.26e-01} & 1.83e-01 (6.14e-02) & 1.64e-01 & 0.99 & 9.34e-02 (3.24e-02) & 8.26e-02 & 0.99 \\
 &  & 10 & 3.64e-01 (1.21e-01) & \multicolumn{2}{c|}{3.26e-01} & 1.83e-01 (6.15e-02) & 1.64e-01 & 0.99 & 9.40e-02 (3.26e-02) & 8.47e-02 & 0.95 \\ \cline{2-12} 
 & \multirow{3}{*}{NIPG} & 1 & 3.75e-01 (1.25e-01) & \multicolumn{2}{c|}{3.38e-01} & 1.86e-01 (6.21e-02) & 1.67e-01 & 1.02 & 9.30e-02 (3.13e-02) & 8.36e-02 & 1.00 \\
 &  & 5 & 3.64e-01 (1.21e-01) & \multicolumn{2}{c|}{3.26e-01} & 1.83e-01 (6.13e-02) & 1.64e-01 & 0.99 & 9.32e-02 (3.21e-02) & 8.29e-02 & 0.99 \\
 &  & 10 & 3.64e-01 (1.21e-01) & \multicolumn{2}{c|}{3.26e-01} & 1.83e-01 (6.17e-02) & 1.64e-01 & 0.99 & 9.34e-02 (3.20e-02) & 8.34e-02 & 0.98 \\ \hline
\multirow{6}{*}{$L^2$} & \multirow{3}{*}{SIPG} & 1 & 9.73e-03 (3.80e-03) & \multicolumn{2}{c|}{9.03e-03} & 2.23e-03 (8.47e-04) & 2.01e-03 & 2.16 & 9.49e-04 (6.29e-04) & 6.85e-04 & 1.55 \\
 &  & 5 & 5.17e-03 (1.70e-03) & \multicolumn{2}{c|}{4.66e-03} & 1.70e-03 (7.46e-04) & 1.46e-03 & 1.66 & 1.05e-03 (8.96e-04) & 7.38e-04 & 0.99 \\
 &  & 10 & 5.43e-03 (1.77e-03) & \multicolumn{2}{c|}{4.98e-03} & 1.75e-03 (8.56e-04) & 1.48e-03 & 1.74 & 1.23e-03 (9.10e-04) & 8.97e-04 & 0.73 \\ \cline{2-12} 
 & \multirow{3}{*}{NIPG} & 1 & 8.99e-03 (4.74e-03) & \multicolumn{2}{c|}{8.12e-03} & 2.56e-03 (1.33e-03) & 2.27e-03 & 1.83 & 1.16e-03 (7.33e-04) & 9.13e-04 & 1.32 \\
 &  & 5 & 4.17e-03 (1.61e-03) & \multicolumn{2}{c|}{3.77e-03} & 1.37e-03 (7.29e-04) & 1.14e-03 & 1.73 & 1.01e-03 (8.09e-04) & 6.72e-04 & 0.76 \\
 &  & 10 & 4.69e-03 (1.67e-03) & \multicolumn{2}{c|}{4.31e-03} & 1.54e-03 (7.26e-04) & 1.33e-03 & 1.70 & 1.02e-03 (7.07e-04) & 7.58e-04 & 0.81 \\ \hline
\end{tabular}%
}
\egroup
\end{table}

\begin{figure}
    \centering
    \includegraphics[width=\textwidth]{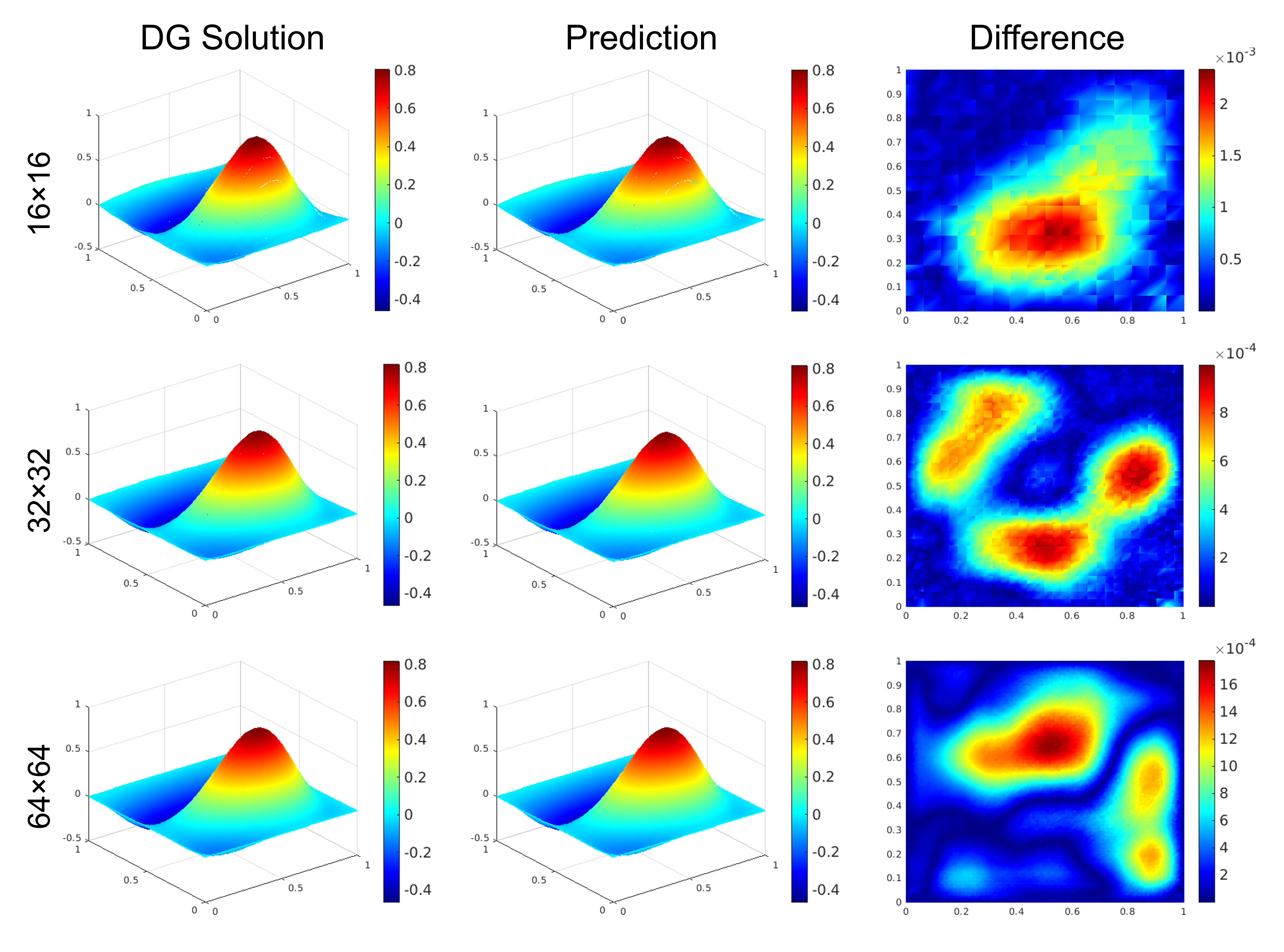}
    \caption{DG solution (left), prediction (center), and absolute difference (right) for SIPG method with $\sigma = 1$ for $N = 16$, $32$, and $64$. \label{fig:sipg_sigma_1_results}}
\end{figure}

Table \ref{tab:results-sup-sigma-n-diff} presents the mean, standard deviation, and median broken $H^1$ and $L^2$ errors between DG solutions and our predicted solutions. The small errors displayed in this table suggest that our supervised approach is learning solutions comparable to those of the DG method. Indeed, Figure \ref{fig:sipg_sigma_1_results} illustrates the DG solutions alongside our predicted solutions and the corresponding differences for various grid sizes, using models trained on data generated from the SIPG method with $\sigma = 1$. The comparison demonstrates that the predictions are visually indistinguishable from the DG solutions across all grid sizes.

\begin{table}[ht!]
\centering
\caption{Mean, standard deviation, and median for the broken $H^1$ and $L^2$ errors between the true and predicted DG solutions for $\varepsilon = -1, +1$ and for varying values of $\sigma$.\label{tab:results-sup-sigma-n-diff}}
\bgroup
\def\arraystretch{1.75}%  1 is the default, change whatever you need
\resizebox{0.95\textwidth}{!}{%
\begin{tabular}{lclccl|cc|cc}
\hline
\multirow{2}{*}{Error} & \multirow{2}{*}{Method} & \multirow{2}{*}{$\sigma$} & \multicolumn{3}{c|}{$N = 16$} & \multicolumn{2}{c|}{$N = 32$} & \multicolumn{2}{c}{$N = 64$} \\ \cline{4-10} 
 &  &  & Mean (Std) & \multicolumn{2}{c|}{Median} & Mean (Std) & Median & Mean (Std) & Median \\ \hline
\multirow{6}{*}{$H^1$} & \multirow{3}{*}{SIPG} & 1 & 8.19e-03 (8.85e-03) & \multicolumn{2}{c|}{5.52e-03} & 4.99e-03 (4.71e-03) & 3.47e-03 & 7.86e-03 (5.81e-03) & 5.81e-03 \\
 &  & 5 & 9.20e-03 (9.39e-03) & \multicolumn{2}{c|}{6.30e-03} & 1.12e-02 (8.58e-03) & 8.42e-03 & 1.61e-02 (1.32e-02) & 1.16e-02 \\
 &  & 10 & 1.28e-02 (1.23e-02) & \multicolumn{2}{c|}{8.61e-03} & 1.37e-02 (1.11e-02) & 1.00e-02 & 1.62e-02 (1.21e-02) & 1.14e-02 \\ \cline{2-10} 
 & \multirow{3}{*}{NIPG} & 1 & 7.05e-03 (8.86e-03) & \multicolumn{2}{c|}{4.40e-03} & 6.52e-03 (5.75e-03) & 4.55e-03 & 1.10e-02 (9.52e-03) & 7.38e-03 \\
 &  & 5 & 1.02e-02 (9.43e-03) & \multicolumn{2}{c|}{7.72e-03} & 1.10e-02 (9.44e-03) & 7.71e-03 & 1.67e-02 (1.25e-02) & 1.21e-02 \\
 &  & 10 & 1.46e-02 (1.59e-02) & \multicolumn{2}{c|}{9.23e-03} & 1.63e-02 (1.19e-02) & 1.15e-02 & 1.68e-02 (1.28e-02) & 1.20e-02 \\ \hline
\multirow{6}{*}{$L^2$} & \multirow{3}{*}{SIPG} & 1 & 8.45e-04 (7.78e-04) & \multicolumn{2}{c|}{6.06e-04} & 5.07e-04 (4.66e-04) & 3.59e-04 & 6.46e-04 (5.41e-04) & 4.69e-04 \\
 &  & 5 & 8.17e-04 (7.67e-04) & \multicolumn{2}{c|}{5.82e-04} & 8.91e-04 (7.10e-04) & 6.62e-04 & 1.02e-03 (8.90e-04) & 6.96e-04 \\
 &  & 10 & 9.58e-04 (8.15e-04) & \multicolumn{2}{c|}{7.37e-04} & 9.34e-04 (8.25e-04) & 6.63e-04 & 9.15e-04 (7.82e-04) & 6.30e-04 \\ \cline{2-10} 
 & \multirow{3}{*}{NIPG} & 1 & 7.16e-04 (7.33e-04) & \multicolumn{2}{c|}{4.71e-04} & 6.69e-04 (5.51e-04) & 5.07e-04 & 9.16e-04 (8.21e-04) & 5.62e-04 \\
 &  & 5 & 8.66e-04 (6.71e-04) & \multicolumn{2}{c|}{6.45e-04} & 8.63e-04 (6.65e-04) & 6.65e-04 & 1.11e-03 (9.18e-04) & 7.46e-04 \\
 &  & 10 & 1.05e-03 (1.08e-03) & \multicolumn{2}{c|}{6.81e-04} & 1.01e-03 (7.25e-04) & 7.21e-04 & 9.25e-04 (8.03e-04) & 6.42e-04 \\ \hline
\end{tabular}%
}
\egroup
\end{table}

\begin{figure}[ht!]
    \centering
    \includegraphics[width=0.3\linewidth]{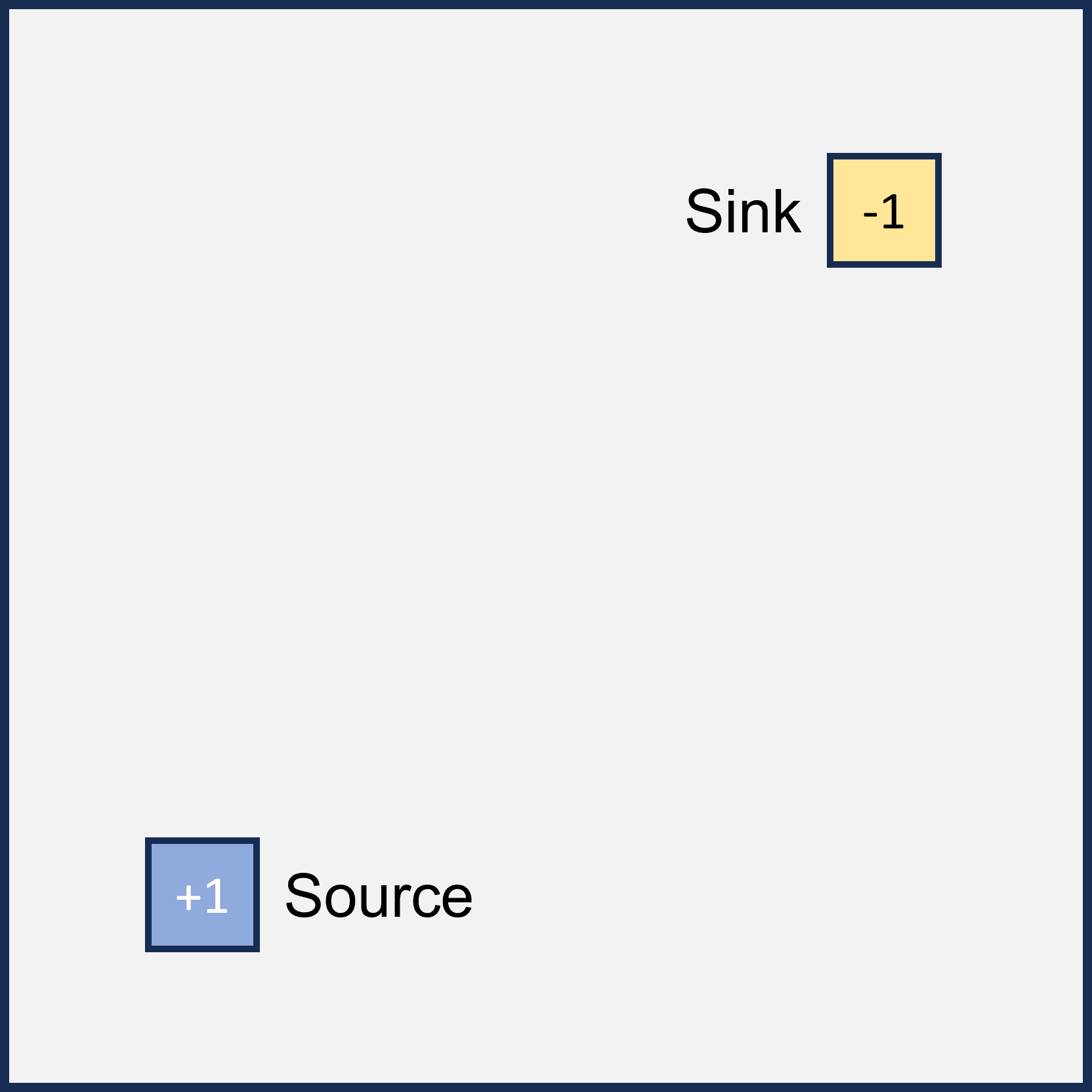}
    \caption{Setup for the function $f$ in the Darcy flow problem. In this case, $f$ is equal to $+1$ in the source region, $-1$ in the sink region, and zero everywhere else.\label{fig:wellschema}}
\end{figure}

\begin{figure}[ht!]
    \centering
    \includegraphics[width=\linewidth]{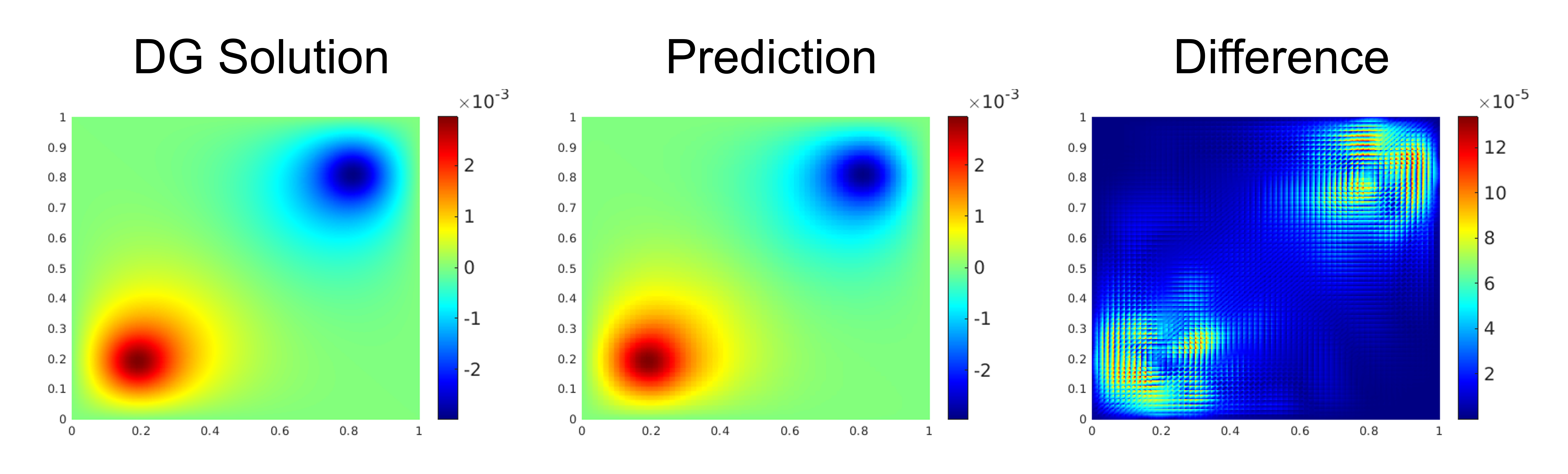}
    \caption{From left to right, DG solution, supervised prediction, and pointwise difference for the Darcy flow problem on a $64\times 64$ grid.\label{fig:supwells}}
\end{figure}

Next, to further test the robustness (generalization) of our supervised approach, we consider the problem of Darcy flow with a source and a sink in the computational domain. 
Let $h=1/64$. The function $f$ is set to zero, except in two small regions:  in the subdomain $(7 h, 17 h) \times(7 h,17 h)$ the function $f$ is a constant equal to $+1$ and in the subdomain $(47 h,57 h)\times(47 h, 57 h)$, it is a constant equal to $-1$ (see Figure~\ref{fig:wellschema}). We do not have an analytical exact solution. The problem models single phase flow from an injection well near the left bottom corner to a production well near the right top corner of $\Omega$.  We apply the trained CNN (with labeled data from $N=64$, $\epsilon = -1$, $\sigma = 1$) to $\pi f$. Figure~\ref{fig:supwells} displays the DG solution, the predicted CNN solution and the pointwise error. We observe that the supervised method produces an accurate solution with an error in the $L^\infty$  norm of the order of $10^{-4}$. 

For the remainder of our discussion on the supervised approach, we will utilize data generated from the SIPG method with $\sigma = 1$ on a 32×32 grid, using the default training parameters outlined in Section \ref{sec:methods-training}. To evaluate the effectiveness of our linear networks, we compare the errors between the DG solutions and the predicted solutions from networks trained with different activation functions. Specifically, we examine the identity activation function and compare it with the ReLU, Tanh, and PReLU activation functions. The results of this experiment are displayed in Table \ref{tab:results-activations-diff}. Our findings indicate that the identity activation function yields the lowest errors in both the broken $H^1$ and $L^2$ norms, suggesting that our choice of linear networks is suitable for learning approximations to DG solutions.

The rationale for not using nonlinear activation functions in our network stems from the nature of the relationship between the $L^2$ projection of the source function $\bfpi_f$ and the degrees of freedom $\bfalpha$. This relationship can be described by the linear system \eqref{eq:linearsystem}. Like with the finite difference-based CNNs in \cite{celaya2024solutions}, we know a priori that our goal is to learn a linear relationship between $\bfpi_f$ and the degrees of freedom. In many machine learning applications, the relationship between input and output is unknown and often assumed to be highly nonlinear. Consequently, incorporating nonlinear activation functions leads to networks that learn nonlinear relationships. By using the identity function as an activation, we move beyond the scope of the universal approximation theorem \cite{Hornik, cybenko1989approximation, funahashi1989approximate}. However, in our case, there is no need for our neural network to serve as a universal approximator; its primary role is to learn a linear mapping.

In addition to using the identity activation function, we omit the use of bias terms in our convolutional layers. We find that the bias terms do not impact the accuracy of our CNN-based solvers and that the values of the bias terms disappear during training. This second point indicates that the bias terms are not necessary for effectively learning DG solutions with CNNs.   

\begin{table}[ht!]
\caption{Mean, standard deviation, and median for the $L^2$ and broken $H^1$ errors between the true and predicted DG solutions for varying activation functions for the test set. %Here, we train each network with data generated from the SIPG method with $\sigma=1$, use 2,000 training examples, set the loss weighting factor to $\beta=0.5$, and use a kernel size of seven. 
The activations resulting in the lowest median errors are highlighted in bold font.\label{tab:results-activations-diff}}
\centering
\bgroup
\def\arraystretch{1.75}%  1 is the default, change whatever you need
\resizebox{0.9\textwidth}{!}{%
\begin{tabular}{lcc|cc}
\hline
\multirow{2}{*}{Activation} & \multicolumn{2}{c|}{$|u_h - \hat{u}_h|_{H^1(\Omega)}$} & \multicolumn{2}{c}{$||u_h - \hat{u}_h||_{L^2(\Omega)}$} \\ \cline{2-5} 
 & Mean (Std) & Median & Mean (Std) & Median \\ \hline
Identity & \textbf{4.99e-03 (4.71e-03)} & \textbf{3.47e-03} & \textbf{5.07e-04 (4.66e-04)} & \textbf{3.59e-04} \\
ReLU & 3.17e-02 (2.18e-02) & 2.59e-02 & 4.53e-03 (4.18e-03) & 3.44e-03 \\
Tanh & 1.42e-01 (8.33e-02) & 1.18e-01 & 1.83e-02 (1.11e-02) & 1.50e-02 \\
PReLU & 6.85e-02 (2.44e-02) & 6.67e-02 & 8.45e-03 (3.79e-03) & 7.84e-03 \\ \hline
\end{tabular}%
}
\egroup
\end{table}

Small CNNs have been shown to be effective for several deep learning and scientific machine learning tasks like medical imaging segmentation and physics-based inverse problems \cite{zhang2024efficient, celaya2024inversion, wang2023semi}. To test the impact of the network size on our supervised approach's ability to recover DG solutions, we vary the number of channels and the kernel size in our CNNs.  Table \ref{tab:results-init-filters} shows the mean, standard deviation, and median errors between the DG and predicted DG solutions in the broken $H^1$ seminorm and $L^2$ norm with the number of channels in our CNN set to 4, 8, 16, 32 (default), and 64. In every case, we see that our supervised approach trained with these networks is capable of accurately approximating DG solutions. In particular, we see that using 32 channels appears to be the optimal choice for the number of channels in our CNN. However, it is worth noting that even the choice of four channels, resulting in a network with roughly 10,000 parameters, can still effectively learn accurate approximations of DG solutions. We see a similar result in Table \ref{tab:results-kernel-sizes}, which shows the effect of the kernel size (with the number of filters set to 32) on the accuracy of our supervised approach. In this case, we see that the choice of 7$\times$7 kernels appears to be optimal, but smaller kernels like 5$\times$5 and 3$\times$ are also capable of learning accurate approximations to DG solutions.

\begin{table}[ht!]
\caption{Mean, standard deviation, and median for the $L^2$ and broken $H^1$ errors between the DG solution $u_h$ and predicted CNN solution $\hat{u}_h$ for varying numbers of initial filters in our network architecture for the test set. Here, we train each network with data generated from the SIPG method with $\sigma=1$, use the identity activation function,  and use a kernel size of seven.\label{tab:results-init-filters}}
\centering
\bgroup
\def\arraystretch{1.75}%  1 is the default, change whatever you need
\resizebox{0.9\textwidth}{!}{%
\begin{tabular}{lccc|cc}
\hline
\multicolumn{1}{c}{\multirow{2}{*}{\begin{tabular}[c]{@{}c@{}}\# Channels\end{tabular}}} & \multirow{2}{*}{\# Parameters} & \multicolumn{2}{c|}{$|u_h - \hat{u}_h|_{H^1(\Omega)}$} & \multicolumn{2}{c}{$||u_h - \hat{u}_h||_{L^2(\Omega)}$} \\ \cline{3-6} 
\multicolumn{1}{c}{} &  & Mean (Std) & Median & Mean (Std) & Median \\ \hline
4 & 1.40e+04 & 3.21e-02 (1.76e-02) & 2.63e-02 & 2.81e-03 (1.33e-03) & 2.46e-03 \\
8 & 5.50e+04 & 2.85e-02 (2.09e-02) & 2.11e-02 & 2.16e-03 (1.15e-03) & 1.85e-03 \\
16 & 2.17e+05 & 1.22e-02 (9.97e-03) & 9.25e-03 & 1.10e-03 (9.59e-04) & 7.82e-04 \\
32 & 8.67e+05 & \textbf{4.99e-03 (4.71e-03)} & \textbf{3.47e-03} & \textbf{5.07e-04 (4.66e-04)} & \textbf{3.59e-04} \\
64 & 3.68e+06 & 6.33e-03 (4.65e-03) & 5.05e-03 & 6.85e-04 (5.15e-04) & 5.56e-04 \\ \hline
\end{tabular}%
}
\egroup
\end{table}

\begin{table}[ht!]
\centering
\caption{Mean, standard deviation, and median for the $L^2$ and broken $H^1$ errors between the DG solution $u_h$ and predicted CNN solution $\hat{u}_h$ for varying kernel sizes in our network architecture for the test set. Here, we train each network with data generated from the SIPG method with $\sigma=1$, use the identity activation function and  $32$ initial filters.\label{tab:results-kernel-sizes}}
\bgroup
\def\arraystretch{1.75}%  1 is the default, change whatever you need
\resizebox{0.9\textwidth}{!}{%
\begin{tabular}{lccc|cc}
\hline
\multicolumn{1}{c}{\multirow{2}{*}{\begin{tabular}[c]{@{}c@{}}Kernel\\ Size\end{tabular}}} & \multirow{2}{*}{\# Parameters} & \multicolumn{2}{c|}{$|u_h - \hat{u}_h|_{H^1(\Omega)}$} & \multicolumn{2}{c}{$||u_h - \hat{u}_h||_{L^2(\Omega)}$} \\ \cline{3-6} 
\multicolumn{1}{c}{} &  & Mean (Std) & Median & Mean (Std) & Median \\ \hline
3 & 2.17e+05 & 1.14e-02 (6.71e-03) & 9.24e-03 & 1.30e-03 (9.77e-04) & 1.04e-03 \\
5 & 5.79e+05 & 7.16e-03 (7.32e-03) & 4.58e-03 & 5.79e-04 (5.83e-04) & 3.68e-04 \\
7 & 8.67e+05 & \textbf{4.99e-03 (4.71e-03)} & \textbf{3.47e-03} & \textbf{5.07e-04 (4.66e-04)} & \textbf{3.59e-04} \\
9 & 1.43e+06 & 6.98e-03 (6.11e-03) & 4.84e-03 & 6.79e-04 (5.41e-04) & 5.25e-04 \\ \hline
\end{tabular}%
}
\egroup
\end{table}

Further modifications can be made to our small networks to reduce the number of parameters even further. For example, future work will consider the use of depthwise separable convolution layers, which have been shown to significantly reduce the parameter count and computational cost while maintaining model performance \cite{howard2017mobilenets, sandler2018mobilenetv2}. These layers separate the spatial convolution and the channel-wise convolution, allowing for a more efficient representation of the network. Furthermore, exploring pruning techniques and knowledge distillation could lead to additional network compression without compromising accuracy. By utilizing these strategies, we may be able to design even smaller CNNs specifically tailored for recovering DG solutions, enhancing the applicability of our approach in resource-constrained environments.

%\Rd I would remove the paragraph below...
%The number of parameters in our neural networks is significantly smaller than the number of entries in $\mathbf{A}$. For example, in the case of a $64 \times 64$ grid, the matrix $\mathbf{A} \in \mathbb{R}^{16,384 \times 16,384}$ contains approximately $2 \times 10^8$ entries. In contrast, our CNNs have between $10^4$ and $10^6$ parameters. Therefore, our CNNs can be viewed as compressed representations of $\mathbf{A}^{-1}$.
%\Bk
%\Rd but how many entries are non-zero? for 64 by 64, this gives at most 81920 non-zero entries in the DG matrix since each element communicates with at most 4 other elements\Bk

\begin{figure}
    \centering
    \includegraphics[width=\linewidth]{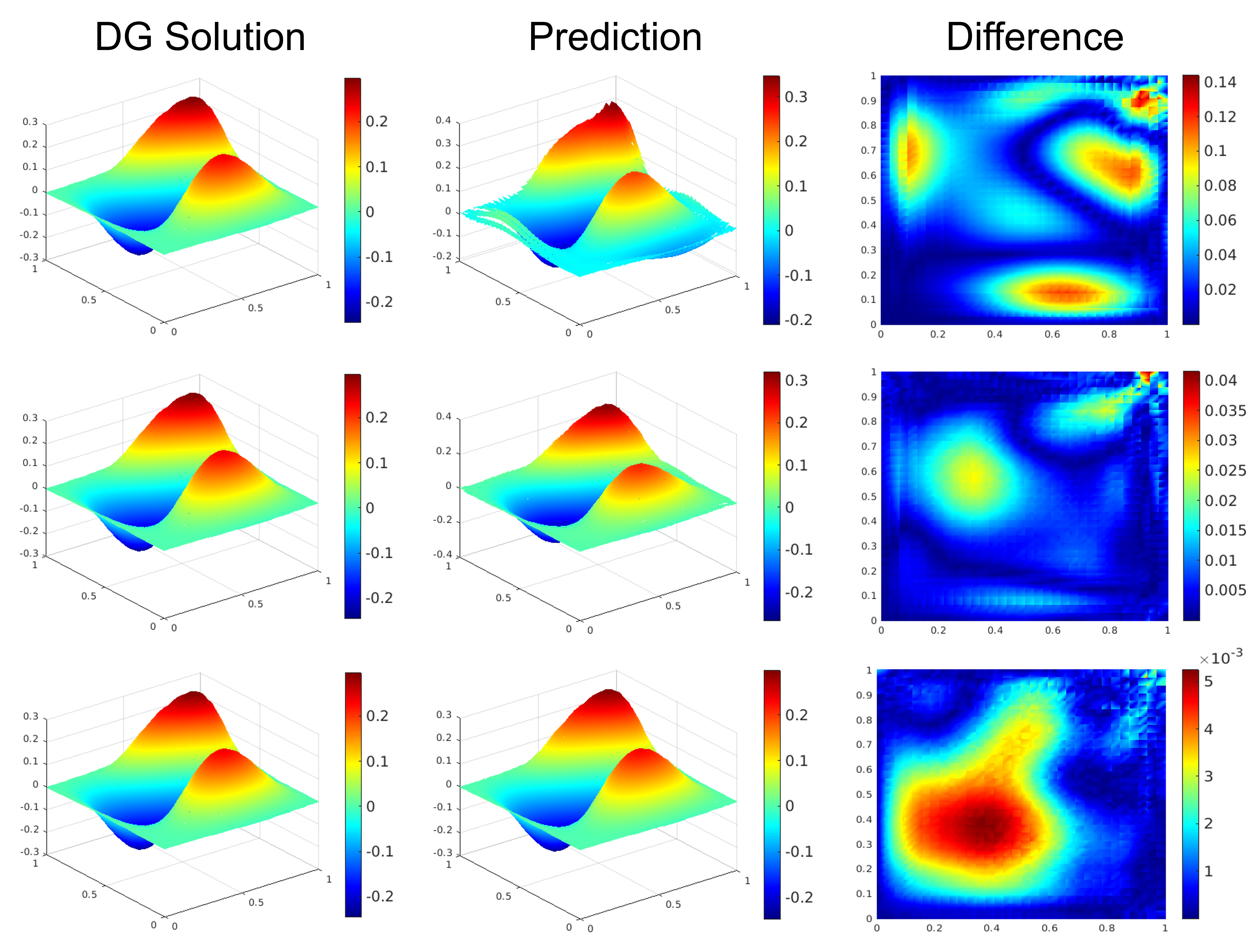}
    \caption{From top to bottom: DG solution, test-time predictions from networks trained with 1, 10, and 100 training examples, and their absolute differences.\label{fig:results-num-train}}
\end{figure}

Next, we examine the influence of training set size on the accuracy of our supervised approach. Table \ref{tab:results-num-train} presents the mean, standard deviation, and median of the broken $H^1$ and $L^2$ errors between the true and predicted DG solutions for networks trained with different sizes of labeled training sets: 1, 10, 100, 1,000, and 2,000 (default) examples. The table clearly demonstrates that accuracy improves as the number of training examples increases, with the most substantial gains occurring between the sets of 1 and 1,000 examples. Figure \ref{fig:results-num-train} displays  test-time predictions from networks trained with 1, 10, and 100 training examples. For the prediction from a network trained with a single example, there are visually noticeable errors. Nonetheless, the model is still able to capture the general shape of the solution. As we increase the number of training examples, we see the scale of the errors decrease by an order of magnitude. This figure and Table \ref{tab:results-num-train} indicate that our supervised approach is able to learn accurate approximations to DG solutions with few training examples. However, the minimum number of training examples required for acceptable results is likely application and problem-dependent. We expect that more complicated, nonlinear, or coupled PDEs will require more data to learn DG solutions effectively.

\begin{table}[ht!]
\centering
\caption{Mean, standard deviation, and median for the $L^2$ and broken $H^1$ errors between the DG solution $u_h$ and predicted DG solution $\hat{u}_h$ for varying training set sizes for the test set. Here, we train each network with data generated from the SIPG method with $\sigma=1$. %, use the identity activation function, set the loss weighting factor to $\beta=0.5$, and use a kernel size of seven.
\label{tab:results-num-train}}
\bgroup
\def\arraystretch{1.75}%  1 is the default, change whatever you need
\resizebox{0.9\textwidth}{!}{%
\begin{tabular}{lcc|cc}
\hline
\multicolumn{1}{c}{\multirow{2}{*}{\begin{tabular}[c]{@{}c@{}}Training Set\\ Size\end{tabular}}} & \multicolumn{2}{c|}{$|u_h - \hat{u}_h|_{H^1(\Omega)}$} & \multicolumn{2}{c}{$||u_h - \hat{u}_h||_{L^2(\Omega)}$} \\ \cline{2-5} 
\multicolumn{1}{c}{} & Mean (Std) & Median & Mean (Std) & Median \\ \hline
1 & 1.16e+00 (5.90e-01) & 9.27e-01 & 1.30e-01 (5.05e-02) & 1.26e-01 \\
10 & 3.09e-01 (2.33e-01) & 2.25e-01 & 2.49e-02 (1.21e-02) & 2.37e-02 \\
100 & 3.55e-02 (2.09e-02) & 2.86e-02 & 3.63e-03 (1.93e-03) & 2.73e-03 \\
1,000 & 8.28e-03 (7.33e-03) & 5.68e-03 & 8.51e-04 (7.86e-04) & 5.61e-04 \\
2,000 & \textbf{4.99e-03 (4.71e-03)} & \textbf{3.47e-03} & \textbf{5.07e-04 (4.66e-04)} & \textbf{3.59e-04} \\ \hline
\end{tabular}%
}
\egroup
\end{table}

Finally, we address the impact of the value of the weighting parameter $\beta$ in our loss function for our supervised approach.  Excluding  the global residual $\Vert \mathbf{A}\hat{\bfalpha}(f) - \bfpi_f \Vert_2$ from our loss results in less accurate predictions. To illustrate this, we train a network with 100 training examples with $\beta = 0$ and various values of $\beta > 0$. In every case where $\beta > 0$, we see errors that are comparable to the result shown in Table \ref{tab:results-num-train} for 100 training examples and $\beta = 0.5$. However, when $\beta = 0$, we see the median $H^1$ error between the DG and predicted solution increase from roughly $2 \times 10^{-2}$ to $1 \times 10^{-1}$. For the $L^2$ error, we still see comparable accuracy in all cases. As we increase the number of training examples, the impact of $\beta$ becomes less pronounced, but the most accurate results come from loss functions that include the global residual.

% \begin{table}[ht!]
% \centering
% \caption{Mean, standard deviation, and median for the $L^2$ and $H^1$ errors between the DG solution $u_h$ and predicted DG solution $\hat{u}_h$ for varying weighting parameters $\beta$ for the loss function for the test set. Here, we train each network with data generated from the SIPG method with $\sigma=1$, using the identity activation function, 32 initial filters, a kernel size of seven, and a grid size of 32.\label{tab:results-beta}}
% \bgroup
% \def\arraystretch{1.75}%  1 is the default, change whatever you need
% \resizebox{0.75\textwidth}{!}{%
% \begin{tabular}{lcc|cc}
% \hline
% \multicolumn{1}{c}{\multirow{2}{*}{$\beta$}} & \multicolumn{2}{c|}{$|u_h - \hat{u}_h|_{H^1(\Omega)}$} & \multicolumn{2}{c}{$||u_h - \hat{u}_h||_{L^2(\Omega)}$} \\ \cline{2-5} 
% \multicolumn{1}{c}{} & Mean (Std) & Median & Mean (Std) & Median \\ \hline
% 0.00 & 3.21e-02 (2.20e-02) & 2.43e-02 & 3.96e-03 (2.33e-03) & 3.38e-03 \\
% 0.10 & 3.73e-02 (2.96e-02) & 2.80e-02 & 4.11e-03 (2.81e-03) & 3.05e-03 \\
% 0.25 & 3.85e-02(2.55e-02) & 3.15e-02 & 4.11e-03 (2.58e-03) & 3.35e-03 \\
% 0.50 & 3.55e-02 (2.09e-02) & 2.86e-02 & 3.63e-03 (1.93e-03) & 2.73e-03 \\
% 0.75 & \textbf{3.22e-02 (2.57e-02)} & \textbf{2.38e-02} & 2.98e-03 (1.17e-03) & 2.55e-03 \\
% 0.90 & 3.44e-02 (2.23e-02) & 2.58e-02 & \textbf{2.37e-03 (1.32e-03)} & \textbf{1.91e-03} \\
% 1.00 & 1.31e-01 (1.06e-01) & 9.32e-02 & 4.87e-03 (3.5e-03) & 3.80e-03 \\ \hline
% \end{tabular}%
% }
% \egroup
% \end{table}

\subsection{Unsupervised approach}
\begin{figure}
    \centering
    \includegraphics[width=\linewidth]{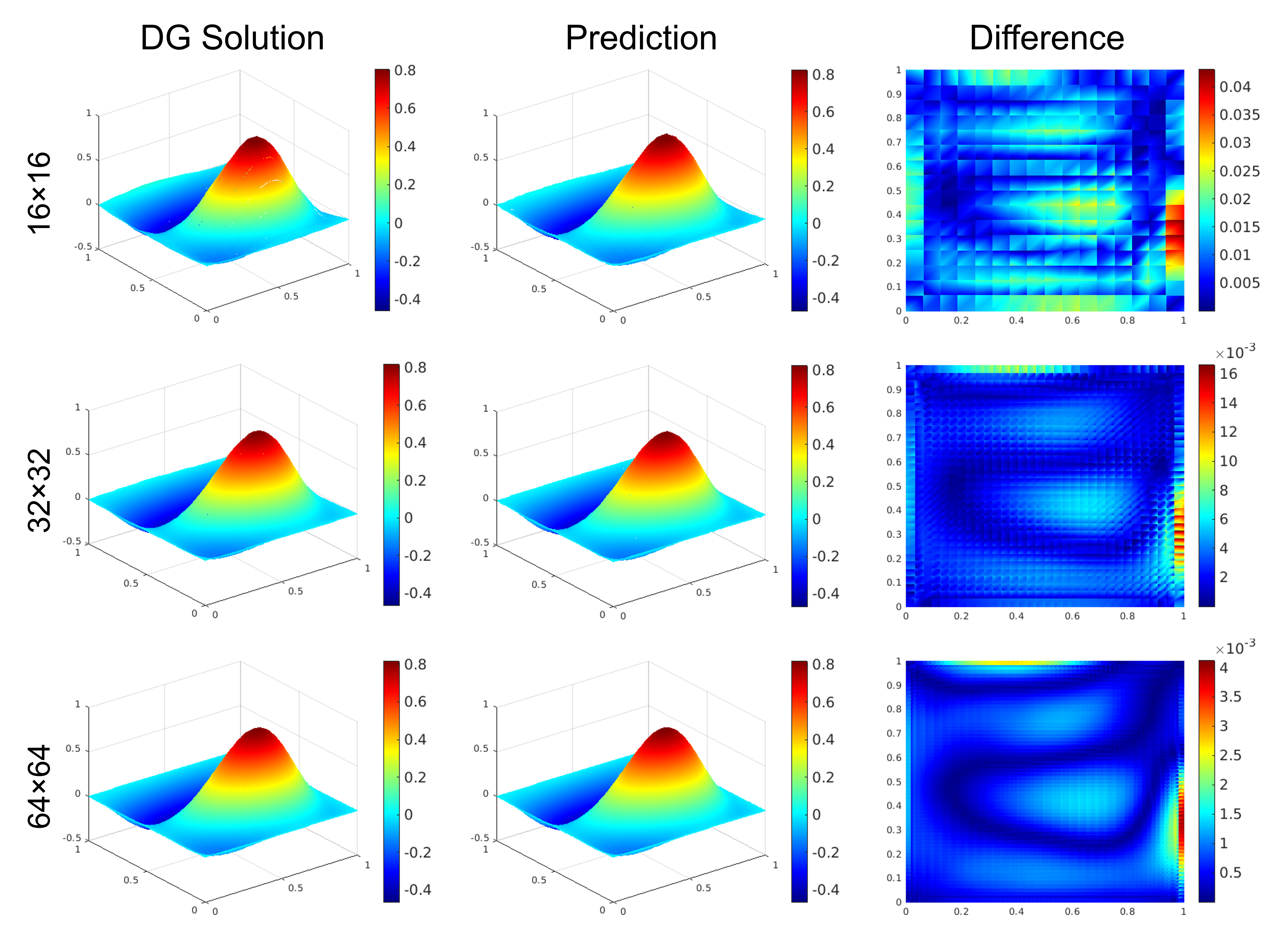}
    \caption{Unsupervised predictions for varying grid sizes. We set $\sigma = 1$ and $\eta = 2$ in the unsupervised loss function and use 5,000 optimization steps.\label{fig:results-unsup-grid-sizes}}
\end{figure}

Our unsupervised approach does not depend on labeled training data to approximate DG solutions, which is a notable advantage over the supervised method. The accuracy of our unsupervised approach is significantly influenced by two parameters: the penalty term $\sigma$ in the local mass error and the discontinuity penalty $\eta$. Table \ref{tab:results-unsup-sigma} presents the mean, standard deviation, median, and rate of convergence for the broken $H^1$ and $L^2$ errors between true solution $u$ and the predicted DG solution $\hat{u}_h$, assessed across varying values of $\sigma$ on the set of $100$ test functions with different grid sizes. In this experiment, we set $\eta = 2$ and use $5,000$ optimization steps. Here, we see that $\sigma=1$ yields the most accurate results, with errors and convergence rates that are similar to those of the DG methods in Table \ref{tab:results-dg-errors}, namely rates equal to $1$ for the $H^1$ error and rates equal to $2$ for the $L^2$ error. Figure \ref{fig:results-unsup-grid-sizes} displays the DG solution (SIPG, $\sigma=1$), the output of our unsupervised approach, and the absolute difference between the two for varying grid sizes. In this figure, we see that our unsupervised approach produces predictions that are visually indistinguishable from the DG solution for each grid size. We also see that the scale of the errors decreases by an order of magnitude as we refine our mesh.

\begin{table}[ht!]
\centering
\caption{ Mean, standard deviation, median, and rate of convergence for the broken $H^1$ and $L^2$ errors between the true solution $u$ and the predicted DG solution $\hat{u}_h$ for varying values of $\sigma$. We fix $\eta = 2$.\label{tab:results-unsup-sigma}}
\bgroup
\def\arraystretch{1.75}%  1 is the default, change whatever you need
\resizebox{\textwidth}{!}{%
\begin{tabular}{llccl|ccc|ccc}
\hline
\multicolumn{1}{c}{\multirow{2}{*}{Error}} & \multirow{2}{*}{$\sigma$} & \multicolumn{3}{c|}{$N = 16$} & \multicolumn{3}{c|}{$N = 32$} & \multicolumn{3}{c}{$N = 64$} \\ \cline{3-11} 
\multicolumn{1}{c}{} &  & Mean (Std) & \multicolumn{2}{c|}{Median} & Mean (Std) & Median & Rate & Mean (Std) & Median & Rate \\ \hline
\multirow{3}{*}{$H^1$} & 1 & \textbf{3.78e-01 (1.38e-01)} & \multicolumn{2}{c|}{\textbf{3.28e-01}} & \textbf{1.90e-01 (6.86e-02)} & \textbf{1.58e-01} & \textbf{1.05} & \textbf{9.03e-02 (3.22e-02)} & \textbf{7.66e-02} & \textbf{1.04} \\
 & 5 & 4.15e-01 (1.44e-01) & \multicolumn{2}{c|}{3.69e-01} & 1.93e-01 (6.48e-02) & 1.67e-01 & 1.14 & 1.19e-01(7.19e-02) & 8.15e-02 & 1.03 \\
 & 10 & 5.11e-01 (1.74e-01) & \multicolumn{2}{c|}{4.24e-01} & 4.38e-01 (2.27e-01) & 4.10e-01 & 0.05 & 4.29e-01 (1.95e-01) & 3.56e-01 & 0.20 \\ \hline
\multirow{3}{*}{$L^2$} & 1 & \textbf{8.02e-03 (4.94e-03)} & \multicolumn{2}{c|}{\textbf{6.36e-03}} & \textbf{2.17e-03 (1.16e-03)} & \textbf{1.71e-03} & \textbf{1.90} & \textbf{5.35e-04 (2.91e-04)} & \textbf{4.22e-04} & \textbf{2.01} \\
 & 5 & 8.83e-03 (3.23e-03) & \multicolumn{2}{c|}{8.18e-03} & 5.44e-03 (6.37e-03) & 2.04e-03 & 1.98 & 5.21e-03 (9.95e-03) & 1.56e-03 & 0.39 \\
 & 10 & 1.98e-02 (8.10e-03) & \multicolumn{2}{c|}{1.93e-02} & 4.34e-02 (2.90e-02) & 3.65e-02 & -0.92 & 5.24e-02 (2.79e-02) & 4.32e-02 & -0.24 \\ \hline
\end{tabular}%
}
\egroup
\end{table}

\begin{figure}[ht!]
    \centering
    \includegraphics[width=\linewidth]{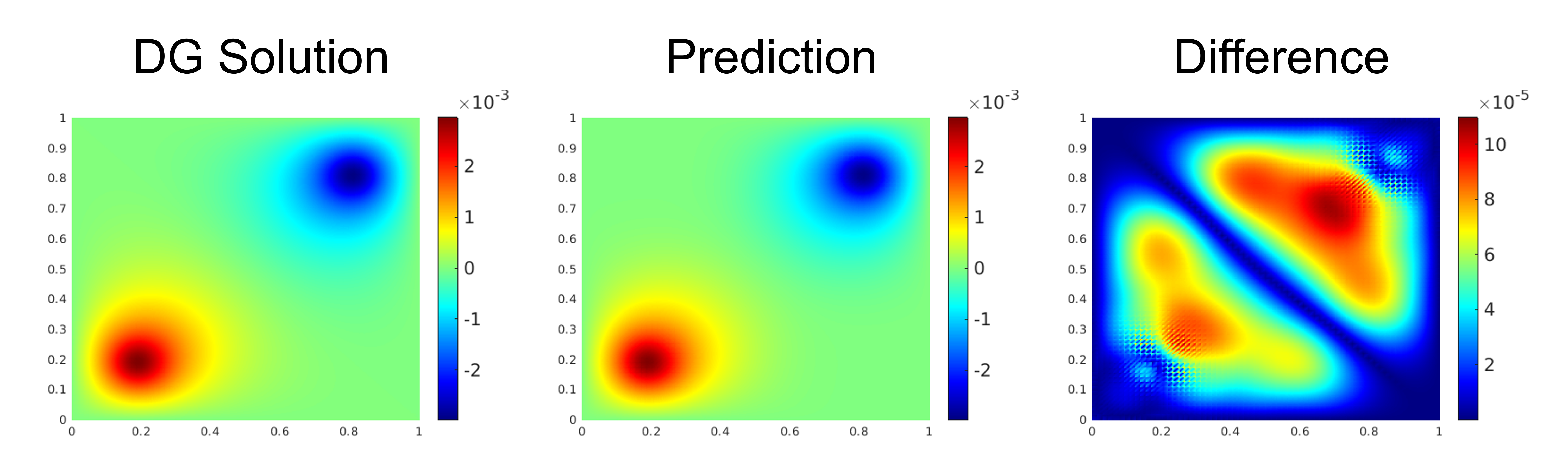}
    \caption{From left to right, DG solution, unsupervised prediction, and pointwise difference for the Darcy flow problem on a $64\times 64$ grid.\label{fig:wellunsup}}
\end{figure}

Next, we apply our unsupervised approach to the source/sink Darcy problem described in the previous section (see Figure~\ref{fig:wellschema}). The input of our CNN is the $L^2$ projection of the function $f$ on a grid $64\times 64$. We choose $\sigma = 1$ and $\eta = 2$ in the loss function and use $5,000$ iterations. Figure~\ref{fig:wellunsup} displays the predicted CNN solution and the pointwise error. As for the supervised case, we observe an accuracy of the order $10^{-4}$ in the $L^\infty$ norm. This example demonstrates the robustness of our approach.

\begin{figure}[ht!]
    \centering
    \includegraphics[width=\linewidth]{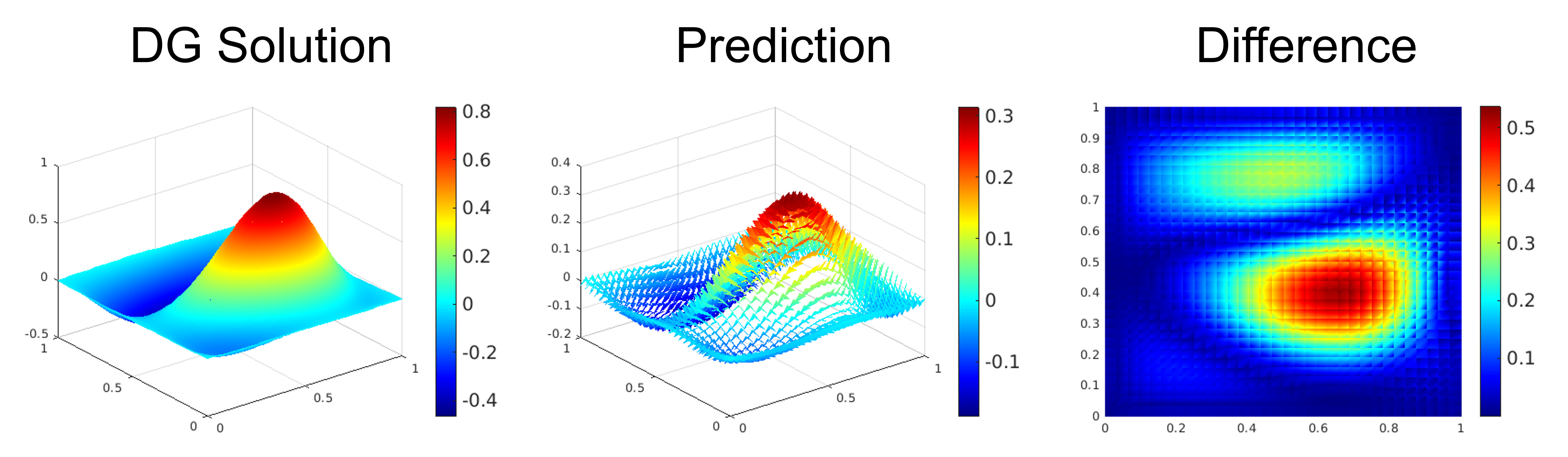}
    \caption{(Left) DG Solution, (Center) Unsupervised prediction with $\eta=0$, and (Right) Difference. \label{fig:results-unsup-eta-zero}}
\end{figure}

For the remainder of the presentation and discussion of our unsupervised approach, we fix a grid size of 32$\times$32 and choose $\sigma=1$. Table \ref{tab:results-unsup-eta} shows the mean, standard deviation, and median of the $L^2$ and broken $H^1$ errors between the true solution and predicted DG solution for varying values of the discontinuity penalty $\eta$. This table shows that for both the broken $H^1$ and $L^2$ errors, the value of $\eta=2$ yields the smallest errors. Additionally, we see that excluding the jump term from the loss function significantly degrades the accuracy of our method. Indeed, for $\eta = 0$, Figure \ref{fig:results-unsup-eta-zero} shows the DG solution (SIPG, $\sigma=1$), the prediction from our unsupervised approach, and their absolute difference. While the unsupervised approach captures the overall shape of the DG solution, it fails to accurately represent the scale and smoothness of the solution. The inclusion of the jump term $\eta\sum_{\gamma\in\Gamma_h} \Vert [\hat u_h]\Vert_{L^2(\gamma)}^2$ is crucial as it helps minimizes the discontinuities between elements.

\begin{table}[ht!]
\centering
\caption{Mean, standard deviation, and median for the broken $H^1$ and $L^2$ errors between the true solution $u$ and the predicted DG solution $\hat{u}_h$ for varying values of $\eta$ in our unsupervised loss.\label{tab:results-unsup-eta}}
\bgroup
\def\arraystretch{1.75}%  1 is the default, change whatever you need
\resizebox{0.9\textwidth}{!}{%
\begin{tabular}{lcc|cc}
\hline
\multicolumn{1}{c}{\multirow{2}{*}{$\eta$}} & \multicolumn{2}{c|}{$|u - \hat{u}_h|_{H^1(\Omega)}$} & \multicolumn{2}{c}{$||u - \hat{u}_h||_{L^2(\Omega)}$} \\ \cline{2-5} 
\multicolumn{1}{c}{} & Mean (Std) & Median & Mean (Std) & Median \\ \hline
0 & 3.08e+00 (7.63e-01) & 2.85e+00 & 1.77e-01 (4.92e-02) & 1.59e-01 \\
1 & 2.19e-01 (9.05e-02) & 1.77e-01 & 2.30e-03 (1.17e-03) & 1.74e-03 \\
2 & \textbf{1.90e-01 (6.86e-02)} & \textbf{1.58e-01} & \textbf{2.17e-03 (1.16e-03)} & \textbf{1.71e-03} \\
5 & 2.12e-01 (7.20e-02) & 1.85e-01 & 2.23e-03 (8.24e-04) & 2.08e-03 \\
10 & 1.96e-01 (7.34e-02) & 1.64e-01 & 4.55e-03 (4.18e-03) & 3.27e-03 \\ \hline
\end{tabular}%
}
\egroup
\end{table}

Another factor in the accuracy of our unsupervised approach is the number of optimization steps. Table \ref{tab:results-unsup-n-steps} shows the mean, standard deviation, and median of the broken $H^1$ and $L^2$ errors between the true solution and predicted DG solution for varying numbers of optimization steps. We fix $\sigma = 1$ and $\eta = 2$ in this case. As anticipated, the errors decrease as the number of optimization steps increases. However, even with a relatively small number of optimization steps (i.e., 500), our unsupervised approach can achieve accuracy comparable to that obtained with 5,000 optimization steps. The number of optimization steps needed to achieve accurate results is likely problem-dependent. While 500 to 5,000 steps work for our model problem, this may not be the case for more complicated PDEs. Future work will investigate this for more complex problems like nonlinear or coupled PDEs.

\begin{table}[ht!]
\centering
\caption{Mean, standard deviation, and median for the broken $H^1$ and $L^2$ errors between the true solution $u$ and the predicted CNN solution $\hat{u}_h$ for varying number of optimization steps.\label{tab:results-unsup-n-steps}}
\bgroup
\def\arraystretch{1.75}%  1 is the default, change whatever you need
\resizebox{0.9\textwidth}{!}{%
\begin{tabular}{lcc|cc}
\hline
\multicolumn{1}{c}{\multirow{2}{*}{\# Steps}} & \multicolumn{2}{c|}{$|u - \hat{u}_h|_{H^1(\Omega)}$} & \multicolumn{2}{c}{$||u - \hat{u}_h||_{L^2(\Omega)}$} \\ \cline{2-5} 
\multicolumn{1}{c}{} & Mean (Std) & Median & Mean (Std) & Median \\ \hline
500 & 4.13e-01(2.74e-01) & 2.79e-01 & 3.88e-03 (2.91e-03) & 2.40e-03 \\
1,000 & 2.99e-01 (1.49e-01) & 2.27e-01 & 2.71e-03 (1.34e-03) & 2.00e-03 \\
2,000 & 2.13e-01 (8.03e-02) & 1.78e-01 & 2.31e-03 (1.29e-03) & 1.79e-03 \\
5,000 & \textbf{1.90e-01 (6.86e-02)} & \textbf{1.58e-01} & \textbf{2.17e-03 (1.16e-03)} & \textbf{1.71e-03} \\ \hline
\end{tabular}%
}
\egroup
\end{table}

Our previous results showed the accuracy of our unsupervised approach compared to the true solution. Table \ref{tab:results-unsup-diff2} shows the mean, standard deviation, and median of the broken $H^1$ and $L^2$ errors between the predicted and DG solutions. In this case, we compute the errors for the SIPG and NIPG methods with different values of $\sigma$ and grid sizes. In this table, we see that our unsupervised approach learns solutions that are comparable to the DG solutions for each configuration of the DG scheme. Notably, we see that our unsupervised approach consistently appears to be learning solutions that are closest to DG solutions from the NIPG method with $\sigma = 1$. We also observe an increase in the convergence rate as we refine the mesh.  This indicates that as the mesh size decreases, the CNN prediction becomes closer and closer to the DG solution.

Like with the supervised approach, the choice of $\beta$ does not significantly impact the accuracy of the unsupervised approach as long as $\beta < 1$. In every case, we see similar results in terms of convergence and the quality of the approximate solution $\hat{u}_h$, indicating that the method is robust to variations in $\beta$.

\begin{table}[ht!]
\centering
\caption{Mean, standard deviation, median and rates for the broken $H^1$ and $L^2$ errors between the predictions from our unsupervised approach and the
DG solutions generated using the SIPG and NIPG methods with different values of $\sigma$. Our unsupervised predictions are generated by setting $\sigma=1$ (in the local mass error loss function), $\eta=2$, and $5,000$ optimization steps.\label{tab:results-unsup-diff2}}
\bgroup
\def\arraystretch{1.75}%  1 is the default, change whatever you need
\resizebox{\textwidth}{!}{%
\begin{tabular}{lclccl|ccc|ccc}
\hline
\multirow{2}{*}{Error} & \multirow{2}{*}{Method} & \multirow{2}{*}{$\sigma$} & \multicolumn{3}{c|}{$N = 16$} & \multicolumn{3}{c|}{$N = 32$} & \multicolumn{3}{c}{$N = 64$} \\ \cline{4-12} 
 &  &  & Mean (Std) & \multicolumn{2}{c|}{Median} & Mean (Std) & Median & Rate & Mean (Std) & Median & Rate\\ \hline
\multirow{6}{*}{$H^1$} & \multirow{3}{*}{SIPG} & 1 & 1.70e-01 (6.09e-02) & \multicolumn{2}{c|}{1.53e-01} & 8.32e-02 (2.86e-02) & 7.67e-02 & 1.03 & 2.00e-02 (9.92e-03) & 1.59e-02 & 2.05\\
 &  & 5 & 1.56e-01 (5.62e-02) & \multicolumn{2}{c|}{1.34e-01} & 7.36e-02 (2.47e-02) & 6.96e-02 & 1.08 & 1.49e-02 (7.79e-03) & 1.18e-02 & 2.30\\
 &  & 10 & 1.59e-01 (5.91e-02) & \multicolumn{2}{c|}{1.38e-01} & 7.39e-02 (2.48e-02) & 6.95e-02 & 1.10 & 1.45e-02 (7.68e-03) & 1.14e-02 & 2.35\\ \cline{2-12} 
 & \multirow{3}{*}{NIPG} & 1 & \textbf{1.01e-01 (3.21e-02)} & \multicolumn{2}{c|}{\textbf{8.91e-02}} & \textbf{5.31e-02 (1.81e-02)} & \textbf{4.75e-02} & 0.93 & \textbf{1.05e-02 (5.67e-03)} & \textbf{8.86e-03} & 2.34\\
 &  & 5 & 1.55e-01 (6.19e-02) & \multicolumn{2}{c|}{1.28e-01} & 6.92e-02 (2.39e-02) & 6.38e-02 & 1.16 & 1.41e-02 (7.33e-03) & 1.07e-02 & 2.29\\
 &  & 10 & 1.59e-01 (6.18e-02) & \multicolumn{2}{c|}{1.33e-01} & 7.19e-02 (2.45e-02) & 6.66e-02 & 1.14 & 1.42e-02 (7.46e-03) & 1.09e-02 & 2.34\\ \hline
\multirow{6}{*}{$L^2$} & \multirow{3}{*}{SIPG} & 1 & 1.44e-02 (5.80e-03) & \multicolumn{2}{c|}{1.14e-02} & 3.73e-03 (1.56e-03) & 3.05e-03 & 1.95 & 8.81e-04 (3.86e-04) & 7.22e-04 & 2.08\\
 &  & 5 & 1.34e-02 (5.37e-03) & \multicolumn{2}{c|}{1.10e-02} & 3.50e-03 (1.41e-03) & 2.90e-03 & 1.94 & 8.41e-04 (3.55e-04) & 6.80e-04 & 2.05\\
 &  & 10 & 1.37e-02 (5.60e-03) & \multicolumn{2}{c|}{1.11e-02} & 3.55e-03 (1.44e-03) & 2.92e-03 & 1.95 & 8.47e-04 (3.60e-04) & 6.86e-04 & 2.06\\ \cline{2-12} 
 & \multirow{3}{*}{NIPG} & 1 & \textbf{7.53e-03 (3.75e-03)} & \multicolumn{2}{c|}{\textbf{5.98e-03}} & \textbf{2.06e-03 (9.33e-04)} & \textbf{1.76e-03} & 1.87 & \textbf{3.67e-04 (2.28e-04)} & \textbf{2.81e-04} & 2.45\\
 &  & 5 & 1.10e-02 (4.19e-03) & \multicolumn{2}{c|}{9.28e-03} & 2.83e-03 (1.02e-03) & 2.49e-03 & 1.96 & 6.30e-04 (2.33e-04) & 5.30e-04 & 2.17\\
 &  & 10 & 1.24e-02 (4.90e-03) & \multicolumn{2}{c|}{1.03e-02} & 3.17e-03 (1.21e-03) & 2.69e-03 & 1.97 & 7.32e-04 (2.89e-04) & 5.97e-04 & 2.11\\ \hline
\end{tabular}%
}
\egroup
\end{table}

\subsection{Computational costs and considerations}
When considering our proposed approaches, it is essential to account for computational costs. With our supervised approach, the biggest bottleneck is data generation. This bottleneck is particularly restrictive as we refine our mesh. However, our results in Table \ref{tab:results-num-train} indicate that achieving accurate results with limited training examples is possible, which can help mitigate the computational cost of generating training data. Training the supervised models is also a bottleneck but is alleviated by our use of small, linear CNNs. Indeed, for 2,000 training examples, a single epoch takes three seconds for data from the 16$\times$16 grid and 20 seconds for the 64$\times$64 grid, resulting in training times of eight and 50 minutes. 

\begin{figure}[ht!]
    \centering
    \includegraphics[width=0.8\linewidth]{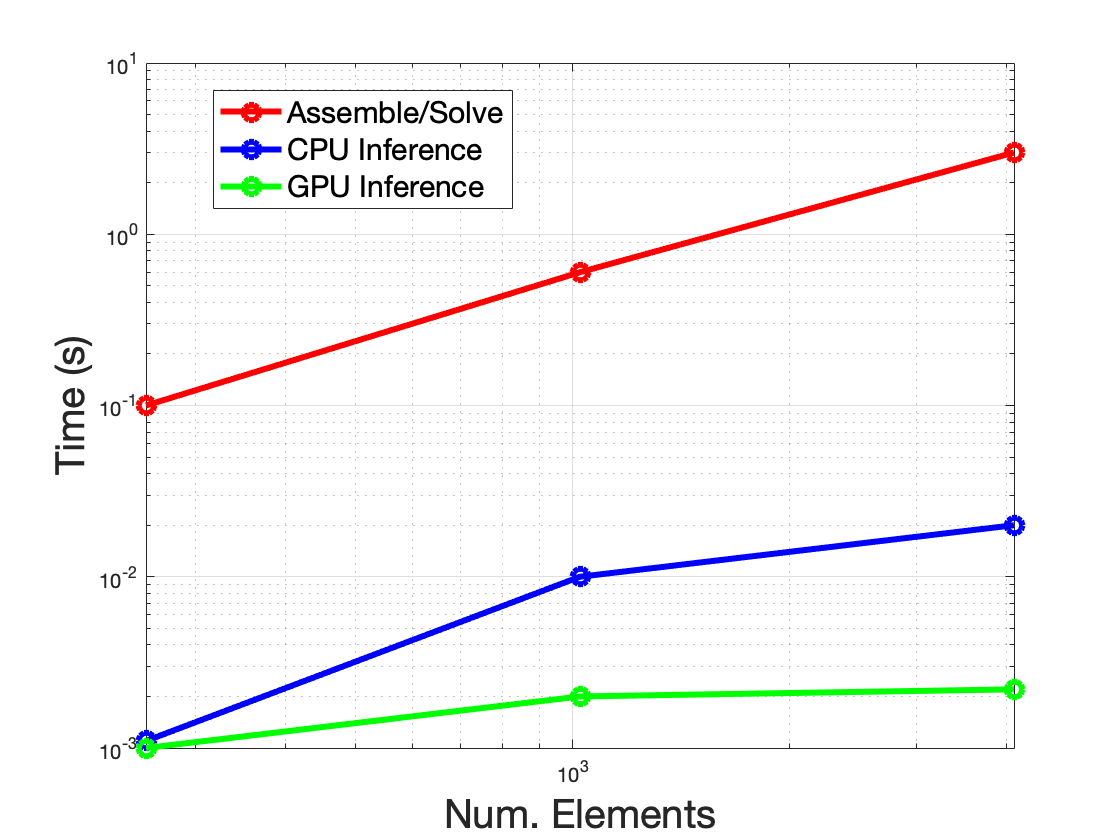}
    \caption{Time to assemble and solve the DG linear system vs. inference with a CPU and GPU for different grid sizes for CNNs trained using our supervised approach.\label{fig:results-inference}}
\end{figure}

Despite these computational bottlenecks for our supervised approach, inference with a trained CNN is fast. Figure \ref{fig:results-inference} compares the inference time for different grid sizes with our supervised approach using a CPU (AMD Ryzen) and GPU (Nvidia A100) with the time required to assemble and solve the linear system \eqref{eq:linearsystem}. In this case, inference with our supervised approach is faster than assembling and solving the linear system. However, neither our implementation of the DG scheme nor PyTorch models are optimized, and further work is needed to conclude how the computational performance of inference with our supervised approach compares to assembling and solving the linear system.

\begin{figure}[ht!]
    \centering
    \includegraphics[width=\linewidth]{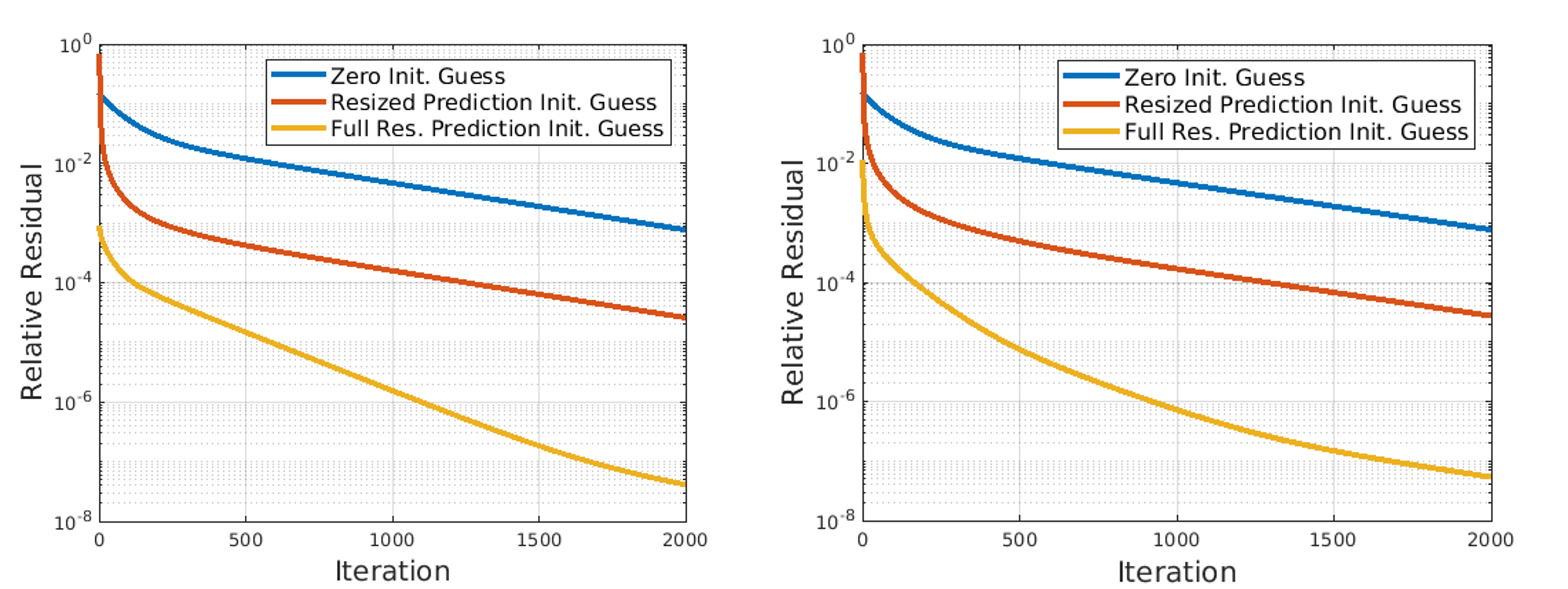}
    \caption{(Left) The relative residual vs. the number of iterations for solving the DG linear system with the Gauss-Seidel method using the following initial guesses: zero, a resampled supervised CNN prediction from a lower resolution grid (16$\times$16), the supervised CNN prediction from the same resolution. This example uses the SIPG method with $\sigma=1$ on a 64$\times$64 grid. (Right) Same as left, but for the unsupervised approach.\label{fig:results-computational-init-guesses}}
\end{figure}

In \cite{celaya2024solutions}, the authors suggested that CNNs based on numerical methods can act as accelerators for solving the resulting linear systems. Here we propose to use our CNN predictions as initial guesses for classical iterative solvers. To test this idea, we compare the convergence of the Gauss-Seidel (GS) method for solving the linear system $\mathbf{A}\bfalpha(f) = \bfpi_f$ resulting from the 64$\times$64 grid for different initial guesses. Specifically, we examine the effect of the following initial guesses: a vector of all zeros, the prediction from our supervised approach trained on the 64$\times$64 grid, and the prediction from our supervised approach trained on a 16$\times$16 grid, which we interpolate (linear interpolation) to the appropriate size for this example linear system. The left side of Figure \ref{fig:results-computational-init-guesses} shows that both of the initial guesses resulting from our supervised approach result in the GS method converging to lower relative residual values in fewer iterations than with a standard zero initial guess. The observation that predictions from lower-resolution models can improve the convergence of numerical solvers for linear systems on finer grids is a benefit of our supervised approach. This result indicates that lower-resolution models, whose data is cheap to generate and fast to train, can provide good initial guesses for numerical solvers on finer grids. 

In contrast to our supervised approach, our unsupervised method does not rely on labeled training data. This eliminates the primary computational bottleneck of the supervised method, which is the generation of labeled data. Moreover, our unsupervised approach avoids the need to build the matrix $\mathbf{A}$, which is a significant computational consideration for DG methods.  When using the same A100 GPU, training our unsupervised method for 5,000 iterations takes between ten and forty seconds on the 16×16 and 64×64 grids, respectively. However, it is important to note that our implementation of the local mass error loss is not yet optimized. Future work will focus on enhancing the computational performance of our unsupervised method. As with the supervised approach, the unsupervised solution can serve as a good initial guess for any iterative solver for the DG linear system. The right side of Figure \ref{fig:results-computational-init-guesses} shows the results of the same experiment with the supervised approach but with predictions from our unsupervised method. In this case, we again see that the full and interpolated low-resolution predictions are better initial guesses for the GS method than the zero initial guess. In both cases, we see that the GS method converges to lower relative residual values in fewer iterations. Additionally, the low-resolution prediction provides a lower-cost alternative to using the unsupervised approach on the higher-resolution grid for generating effective initial guesses for numerical iterative solvers like the GS method. 

The choice between the supervised and unsupervised approaches depends on the specific problem requirements and computational resources available to users. The supervised approach is particularly suitable when labeled training data is readily available or when the computational cost of generating such data is acceptable. This method is especially advantageous when working with lower-resolution grids, as training and inference times are significantly reduced compared to higher-resolution grids, and predictions from these models can be upsampled to provide good initial guesses for linear systems from data on higher-resolution grids. Moreover, the supervised approach may be beneficial in applications where similar problems need to be solved repeatedly on a fixed grid, such as parameterized PDEs. In these cases, once a model is trained, its fast inference can significantly reduce computational costs over multiple problem instances. However, for problems where labeled data generation is prohibitively expensive or impractical, the unsupervised approach may be a good alternative. By leveraging the local mass error loss function, the unsupervised method circumvents the need for labeled data entirely. It also eliminates the computational expense of constructing the matrix $\mathbf{A}$ in DG methods. Thus, our unsupervised approach may be better suited for large-scale problems than its supervised counterpart, particularly in scenarios where fine-grid solutions are required and labeled data is not readily available. Both approaches, however, demonstrate their utility in improving the convergence of numerical solvers, making them valuable tools for DG methods.

\section{Conclusions}
The results presented above show the effectiveness of our proposed supervised and unsupervised approaches for learning approximations of DG solutions to elliptic problems. Unlike classical PINNs, our approach is influenced by numerical PDEs (i.e., the DG method), resulting in more explainable solutions. For both of our proposed approaches, we see that they are not only capable of producing accurate approximations of DG solutions but that these predictions are also good initial guesses for iterative solvers for the linear systems resulting from the DG discretization. As a result, our methods have the potential to reduce the computational cost of solving such systems. 

For applications that require solving the same or similar problems, such as parametric PDEs, the supervised approach may be advantageous. In these cases, users pay the upfront computational cost of generating labeled data and training the model, but the trained model can provide fast predictions. On the other hand, the unsupervised approach is likely better suited for scenarios where labeled training data is unavailable or impractical to generate, leveraging the local mass error loss to provide accurate solutions without training data or assembling a linear system.

Both approaches demonstrate that leveraging insights from traditional numerical methods for PDEs with machine learning can lead to efficient, explainable, and practical solutions to PDEs. Future work will focus on extending these methods to more complex PDEs, optimizing their computational performance, and extending these methods to unstructured grids.

\bibliographystyle{siamplain}
\bibliography{references}
\end{document}

%% file: shared.tex
\usepackage{lipsum}
\usepackage{amsfonts}
\usepackage{graphicx}
\usepackage{epstopdf}
\usepackage{algorithmic}
\usepackage{todonotes}
\usepackage{multirow}
\ifpdf
  \DeclareGraphicsExtensions{.eps,.pdf,.png,.jpg}
\else
  \DeclareGraphicsExtensions{.eps}
\fi

\newcommand{\bfalpha}{\boldsymbol{\alpha}}
\newcommand{\bfpi}{\boldsymbol{\pi}}

\newsiamremark{remark}{Remark}
\newsiamremark{hypothesis}{Hypothesis}
\crefname{hypothesis}{Hypothesis}{Hypotheses}
\newsiamthm{claim}{Claim}

\headers{DG Solutions via Small Linear CNNs}{A. Celaya, Y. Wang, D. Fuentes, and B. Riviere}

\title{Learning Discontinuous Galerkin Solutions to Elliptic Problems via Small Linear Convolutional Neural Networks\thanks{Submitted to the editors February 12, 2025.
\funding{The Department of Defense supports Adrian Celaya through the National Defense Science \& Engineering Graduate Fellowship Program. David Fuentes is partially supported by R21CA249373, NSF-DMS 2111147 and NSF-DMS 2111459; Beatrice Riviere  by NSF-DMS 2111459. This research was partially supported by the Tumor Measurement Initiative through the MD Anderson Strategic Research Initiative Development (STRIDE).}}}

\author{Adrian Celaya\thanks{Dept. of Computational Applied Mathematics and Operations Research, Rice University}
\and Yimo Wang\footnotemark[2]
\and David Fuentes\thanks{Dept. of Imaging Physics, The University of Texas MD Anderson Cancer Center}
\and Beatrice Riviere\thanks{Ken Kennedy Institute, Dept. of Computational Applied Mathematics and Operations Research, Rice University}}

\usepackage{amsopn}